%% file: main.tex
\documentclass[mnsc,nonblindrev]{researchgate}

\usepackage{algorithm,algorithmicx}
\usepackage[noend]{algpseudocode}
\usepackage{xspace}
\usepackage{url, hyperref}
\usepackage{multirow, array}
\usepackage{multicol, graphicx}
\usepackage[flushleft]{threeparttable}
\usepackage{diagbox}

\usepackage{tikz}

\OneAndAHalfSpacedXI

\usepackage{natbib}
 \bibpunct[, ]{(}{)}{,}{a}{}{,}%
 \def\bibfont{\small}%
 \usepackage{enumitem}
 
\usepackage{booktabs}
\TheoremsNumberedThrough     
\ECRepeatTheorems

\EquationsNumberedThrough    

\input{abbrv_su_2023.tex}

\MANUSCRIPTNO{MS-RMA-20-xxxxx}

\begin{document}
\RUNAUTHOR{Jia et al.}
\RUNTITLE{Markdown Pricing Under Unknown Parametric Demand}

\TITLE{Markdown Pricing Under an Unknown Parametric Demand Model}
\ABSTRACT{Consider a single-product revenue-maximization problem where the seller {\em monotonically} decreases the price in $n$ rounds with an unknown demand model coming from a given family.
Without monotonicity, the minimax regret is $\tilde O(n^{2/3})$ for the Lipschitz demand family \citep{kleinberg2003value} and $\tilde O(n^{1/2})$ for a general class of parametric demand models \citep{broder2012dynamic}.
With monotonicity, the minimax regret is $\tilde O(n^{3/4})$ if the revenue function is Lipschitz and unimodal \citep{chen2021multi,jia2021markdown}.
However, the minimax regret for parametric families remained open.
In this work, we provide a complete settlement for this fundamental problem.
We introduce the {\it crossing number} to measure the complexity of a family of demand functions.
In particular, the family of degree-$k$ polynomials has a crossing number $k$. 
Based on {\it conservatism under uncertainty}, we present (i) a policy with an optimal $\Theta(\log^2 n)$ regret for families with crossing number $k=0$, and (ii) another policy with an optimal $\tilde \Theta(n^{k/(k+1)})$ regret when $k\ge 1$.
These bounds are \asymly\ higher than the $\tilde O(\log n)$ and $\tilde \Theta(\sqrt n)$ minimax regret for the same families without the monotonicity constraint \citep{broder2012dynamic}.}


\ARTICLEAUTHORS{%
\AUTHOR{Su Jia}
\AFF{Center of Data Science for Enterprise and Society, Cornell University} 
\AUTHOR{Andrew A. Li, R. Ravi}
\AFF{Tepper School of Business, Carnegie Mellon University} }


\maketitle

\section{Introduction}\label{sec:intro}

\input{intro.tex}

\section{Formulation and Basic Assumptions}
\label{sec:formulation}
\input{model.tex}

\section{Crossing Number}
\label{sec:crossing}

\input{crossing.tex}

\input{results_and_outline.tex}

\bibliographystyle{informs2014}
\bibliography{main.bib}

\newpage
\begin{APPENDICES}
\section{Omitted Proofs in Section \ref{sec:formulation}}
\label{apdx:model}
\input{apdx_prelim.tex}

\section{Details of the Upper Bounds}
\label{apdx:ub}
\input{ub.tex}

\section{Details of the  Lower Bounds}
\label{apdx:lb}
\input{lb.tex}

\end{APPENDICES}
\end{document}

%% file: abbrv_su_2023.tex
\usepackage{pifont}

\newtheorem{observation}{Observation}

\newcommand{\clean}{\color{black}}

\newcommand{\bitem}{\begin{itemize}}
\newcommand{\eitem}{\end{itemize}}
\newcommand{\benum}{\begin{enumerate}}
\newcommand{\eenum}{\end{enumerate}}
\newcommand{\bdefn}{\begin{definition}}
\newcommand{\edefn}{\end{definition}}
\newcommand{\bprop}{\begin{proposition}}
\newcommand{\eprop}{\end{proposition}}
\newcommand{\bque}{\begin{question}}
\newcommand{\eque}{\end{question}}
\newcommand{\bobsv}{\begin{observation}}
\newcommand{\eobsv}{\end{observation}}
\newcommand{\beqn}{\begin{equation}\begin{aligned}}
\newcommand{\eeqn}{\end{aligned}\end{equation}}
\newcommand{\ps}{\begin{proof}[Sketch]}
\newcommand{\brmk}{\begin{remark}}
\newcommand{\ermk}{\end{remark}}
\newcommand{\bduiqi}{\begin{aligned}}
\newcommand{\eduiqi}{\end{aligned}}
\newcommand{\bcoro}{\begin{corollary}}
\newcommand{\ecoro}{\end{corollary}}
\newcommand{\bcom}{}

\newcommand{\asym}{asymptotic}

\newcommand{\asymly}{asymptotically}

\newcommand{\adap}{adaptive}

\newcommand{\arb}{arbitrary}

\newcommand{\Alg}{Algorithm}

\newcommand{\assu}{assumption}

\newcommand{\cond}{condition}
\newcommand{\conf}{confiden}
\newcommand{\confi}{confidence}

\newcommand{\clf}{classifier}

\newcommand{\constr}{constraint}

\newcommand{\contra}{contradiction}

\newcommand{\corres}{corresponding}

\newcommand{\ci}{confidence interval}

\newcommand{\distr}{distribution}

\newcommand{\dec}{decision}

\newcommand{\dxs}{polynomial}

\newcommand{\emp}{empirical}

\newcommand{\elem}{element}

\newcommand{\func}{function}

\newcommand{\ho}{\mathbb}

\newcommand{\indep}{independent}

\newcommand{\IOW}{In other words}

\newcommand{\ih}{induction hypothesis}

\newcommand{\ins}{instance}

\newcommand{\ineq}{inequality}
\newcommand{\ineqs}{inequalities}
\newcommand{\IFT}{It follows that}

\newcommand{\ift}{it follows that}

\newcommand{\lb}{\left}

\newcommand{\lar}{\leftarrow}

\newcommand{\mtx}{matrix}

\newcommand{\ow}{otherwise}
\newcommand{\omg}{\omega}
\newcommand{\Omg}{\Omega}

\newcommand{\OTOH}{On the other hand}

\newcommand{\prb}{probability}

\newcommand{\parti}{particular}

\newcommand{\pmtn}{parametrization}
\newcommand{\pmt}{parameter}

\newcommand{\rv}{random variable}
\newcommand{\rb}{\right}

\newcommand{\rand}{random}
\newcommand{\rar}{\rightarrow}

\newcommand{\real}{\mathbb{R}}
\newcommand{\resp}{respectively}

\newcommand{\sat}{satisfy}
\newcommand{\sats}{satisfies}
\newcommand{\satd}{satisfied}

\newcommand{\sep}{separat}

\newcommand{\sln}{solution}
\newcommand{\sps}{suppose}
\newcommand{\Sps}{Suppose}

\newcommand{\strfwd}{straightforward}

\newcommand{\unk}{unknown}

\newcommand{\wh}{\widehat}

\newcommand{\xulie}{sequence}

\let\eps\varepsilon

%% file: intro.tex
A core challenge in revenue management involves implementing dynamic pricing in the face of uncertain demand, particularly when introducing new products. 
A critical aspect overlooked in the current literature is the {\em direction} of price changes.
Traditional models may involve sellers strategically raising prices to maximize earnings per unit, a strategy considered reasonable when the anticipated reduction in demand is relatively low.
However, in practice, there is a disparity in the treatment of price increases and price decreases ({\em markdowns}). 
This is because the negative impact posed by price increases cannot always be offset by the additional per-unit gain. 
For example, price increases may contribute to a perception of manipulation and adversely affect the seller's ratings.
More concretely, through an analysis of the online menu prices of a group of restaurants, \cite{luca2021effect} found that
\begin{center}
\it  ``a price increase of 1\% leads to a decrease of 3\% – 5\% in the average rating.''
\end{center}

As a result, sellers may incur an implicit penalty for implementing price increases. 
We study the fundamental aspects of this directional asymmetry in price changes by imposing a {\em monotonicity} constraint. 
The objective is to identify a {\em markdown policy} - a policy that never increases the price - to maximize revenue in the presence of an unknown demand model.

Despite its practical importance, markdown pricing under demand uncertainty is less well understood, compared to the unconstrained pricing problem (i.e., without the monotonicity constraint).
Existing work only focused on nonparametric models.
\cite{chen2021multi} and \cite{jia2021markdown} showed that unimodality and Lipschitzness in the {\it revenue} \func\ (i.e., the product of price and mean demand) are sufficient and necessary to achieve sublinear regret.
They proposed a policy with an optimal $\tilde O(n^{3/4})$ regret, i.e., loss due to not knowing the true model.
In addition, if the demand function is twice continuously differentiable, the regret can be improved to $\tilde O(n^{5/7})$; see Theorem 2 of \citealt{jia2021markdown}.

However, in practice, the demand function often exhibits specific structures. 
For example, it is usually reasonable to assume in various applications that the mean demand decreases linearly in the price. 
This added structure enables the seller to learn the model more efficiently, consequently suggesting stronger regret bounds.
This motivates our first question: For parametric demand families, can we find a markdown policy that outperforms those designed for nonparametric families? Formally, 
\begin{center} \it 
Q1) Can we obtain an $o(n^{5/7})$ regret using  a markdown policy for parametric families?
\end{center}


Orthogonal to obtaining a better performance guarantee, we are also interested in quantifying the impact of imposing the monotonicity constraint.
Intuitively, monotonicity complicates the learning-vs-earning trade-off, so it is reasonable to expect the minimax regret of markdown policies to be higher than that of unconstrained policies. 
Such ``separation'' results are known for nonparametric families. 
On the one hand, any markdown policy has $\Omg(n^{3/4})$ regret if the revenue function is Lipschitz and unimodal; see Theorem 3 of \citealt{chen2021multi}.
On the other hand, without the monotonicity constraint, an $\tilde O(n^{2/3})$ regret can be achieved for the same family; see Theorem 3.1 in \citealt{kleinberg2005nearly}.

This leads to our next question: Is there a {\em separation} between markdown and unconstrained pricing for parametric families? 
To formalize this question, note that for parametric families, an  $\tilde O(n^{1/2})$ regret can be achieved for a fairly general class of demand families; see Theorem 3.6 in \citealt{broder2012dynamic}.
Therefore, our question can be formalized as: 
\begin{center} \it 
Q2) Can we show that every markdown policy has an $\omg(\sqrt n)$ regret on some parametric family?
\end{center}

In this work, we provide affirmative answers to these two questions. 
To achieve this, we introduce the {\it crossing number} to measure the complexity of a family of demand models. 
Within this framework, we present minimax optimal regret bounds for every finite crossing number, thereby fully resolving the problem.

\subsection{Our Contributions.}


Our contributions can be classified into three main categories.

\begin{enumerate}
    \item {\bf The Crossing Number.} A central goal in statistical learning involves identifying a suitable metric for the complexity of a model family and characterizing the minimax performance guarantee based on this metric. 
    We introduce the {\it crossing number} that quantifies the maximum number of intersections between any two demand curves. 
    This framework accommodates many existing results in {\em unconstrained} pricing.
    For example, minimax regret is $\Theta(\log n)$ for families for crossing number $k=0$ and $\tilde \Theta(\sqrt n)$ for $1\le k<\infty$; see \citealt{broder2012dynamic}.
    \item {\bf Novel Markdown Policies.} Our contribution at the algorithmic level involves the introduction of two markdown policies, grounded in the principle of {\em conservatism under uncertainty}.
    \benum
    \item {\bf The Cautious Myopic (CM) policy.} The CM policy partitions the time horizon into $O(\log n)$ phases. 
    In each phase, we select a fixed price sufficiently many times and obtain a \ci\ for the optimal price.
    Then we move to the {\em right} endpoint of the \ci\ (as long as monotonicity is preserved).
    We show that the CM policy has $O(\log^2 n)$ regret on any family with crossing number $0$; see Theorem \ref{thm:ub0d_s}.
    \item {\bf The Iterative Cautious Myopic (ICM) policy.} The ICM policy handles crossing number $k\ge 1$. 
    In each phase, we select $(k+1)$ {\em exploration prices} and obtain a \ci\ for the optimal price. 
    As the key difference from the case $k=0$, we decide the prices for the next phase based on the relationship between the \ci\ and the current price. 
    The exploration phase ends if there is sufficient evidence that the current price is lower than the optimal price.
    The ICM policy achieves $\tilde O(n^{k/(k+1)})$ regret on any family with crossing number $k$; see Theorem \ref{thm:ubgen}.
    \item {\bf Improving existing regret bounds.} Prior to this work, the best known regret bound for non-parametric, twice continuously differentiable family is $\tilde O(n^{5/7})$; see Theorem 2 in \citealt{jia2021markdown}.
    Our results improve the above bound (i) to $O(\log^2 n)$ for $k=0$, (ii) to $\tilde O(\sqrt n)$ for $k=1$ and (iii) to $\tilde O(n^{2/3})$ for $k=2$, thereby addressing Q1.
    \eenum
    \item {\bf Tight Lower Bounds.}
    We complement our upper bounds with a matching lower bound for each crossing number $k\ge 0$. 
    These results not only establish the optimality of our upper bounds, but also provide separation from unconstrained pricing, addressing Q2.
    \benum 
    \item {\bf Zero-crossing family.} We show that every markdown policy has $\Omg(\log^2 n)$ regret on a family with crossing number $0$; see Theorem \ref{thm:lb0d}. 
    This lower bound is higher than the $O(\log n)$ regret bound {\em without} the monotonicity \constr\ (see Theorem 4.8 in \citealt{broder2012dynamic}).
    \item {\bf Finite-crossing family.} For any finite $k\ge 1$, we show that any markdown policy has an $\Omg(n^{k/(k+1)})$ regret on a family with crossing number $k$.
    This bound is higher than the $\tilde O(n^{1/2})$ regret bound {\em without} the monotonicity \constr\ (see Theorem 3.6 in  \citealt{broder2012dynamic}).
    \eenum
\end{enumerate}

We summarize our contributions by comparing our results with known regret bounds in Table~\ref{tab:nov3}.


\begin{table}[h]
\centering
\begin{tabular}{|c|c|c|c|c|}
\hline
Crossing number & Unconstrained Pricing  & Markdown Pricing  \\ \hline
$k=0$ & $\Theta(\log n)$ & {\bf \color{red}$\Theta(\log^2 n)$} \\ \hline
$1\le k<\infty$ & $\tilde \Theta(\sqrt n)$ & $\color{red}\tilde 
\Theta(n^{k/(k+1)})$ 
\\ \hline
\end{tabular}
\caption{Minimax regret in terms of the crossing number. 
We highlight our new results in red. 
We emphasize that each entry corresponds to two theorems, an upper bound and a matching lower bound.}
\label{tab:nov3}
\end{table}

\subsection{Related Work}
We summarize previous literature along two dimensions.
In methodology our work is most related to the {\em Multi-Armed Bandit} (MAB) problem. 
In applications, our work is related to the broad literature  of revenue management, especially markdown pricing and new-product pricing.

Dynamic pricing under \unk\ demand model is often cast as an MAB problem given a finite set of arms  (corresponding to the prices), with each providing a revenue drawn i.i.d. from an \unk\ probability distribution specific to that arm.
The objective of the player is to maximize the total revenue earned by pulling a sequence of arms \citep{lai1985asymptotically}.
Our pricing problem generalizes this framework by using an infinite action space $[0,1]$ with each price $p$ corresponding to an action whose revenue is drawn from an \unk\ \distr\ with mean $R(p)$.
In the \textit{Lipschitz Bandits} problem (see, e.g., \cite{agrawal1995continuum}), it is assumed that each $x\in [0,1]$ corresponds to an arm with mean reward $\mu(x)$, where $\mu$ is an \unk\  $L$-Lipschitz \func.
For the one-crossing case, \cite{kleinberg2005nearly} showed a tight $\tilde \Theta(n^{2/3})$ regret bound. 

\begin{figure}[h]
\centering
\includegraphics[width=16cm]{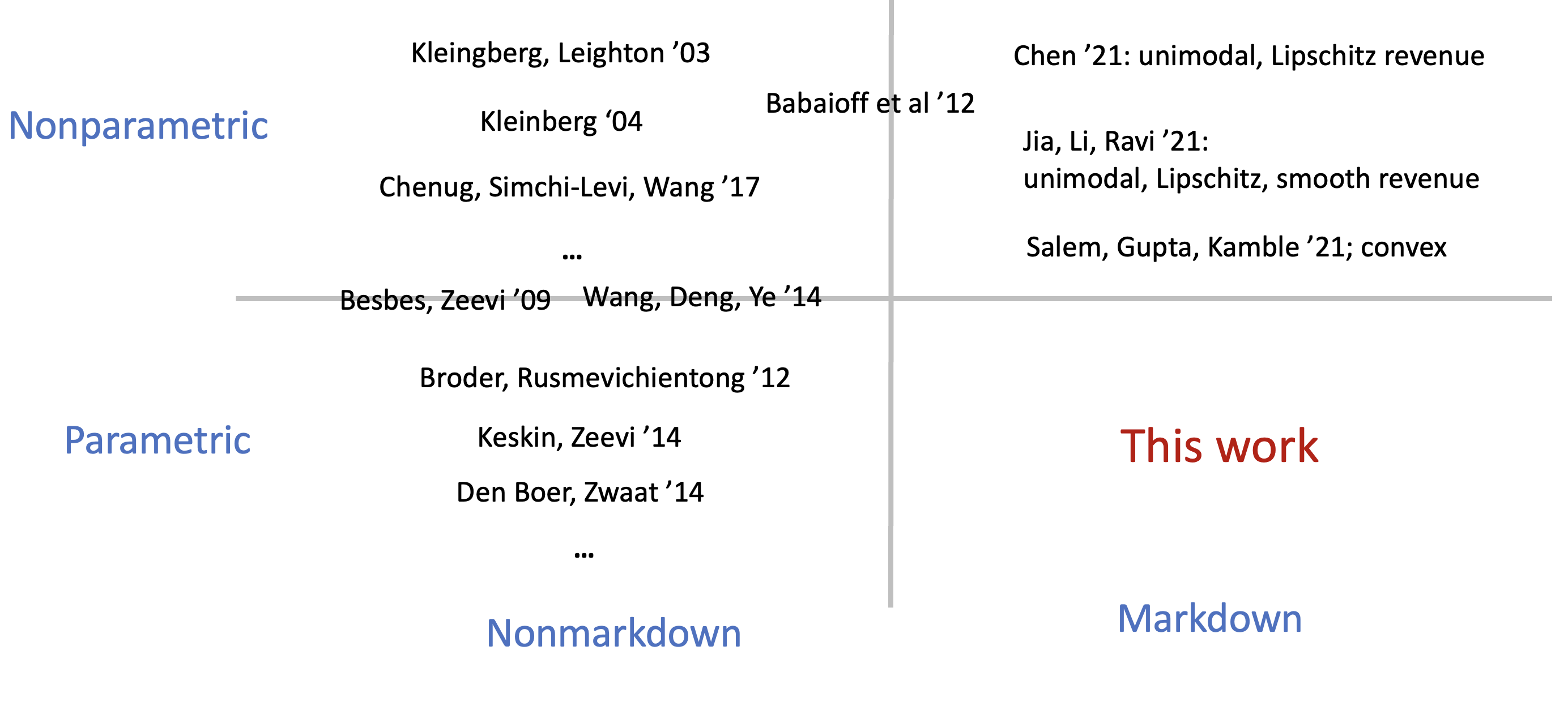}
\caption{Landscape of dynamic pricing with {\color{red} unknown} demand model. 
Papers that fit into two quadrants (e.g., \citealt{besbes2009dynamic}, which involved both parametric and nonparametric models) are placed on the boundary.}
\label{fig:good_dangerous_bad}
\end{figure}

Recently there is an emerging line of work on bandits with monotonicity \constr. 
\cite{chen2021multi}  and \cite{jia2021markdown} recently independently considered the dynamic pricing problem with monotonicity \constr\ under \unk\ demand \func, and proved nearly-optimal $n^{3/4}$ regret bound assuming the reward \func\ is Lipschitz and unimodal.
Moreover, they also showed that these \assu s are indeed minimal, in the sense that no markdown policy achieves $o(T)$ regret without any one of these two \assu s.
Motivated by fairness constraints, \cite{gupta2019individual} and \cite{salem2021taming} considered a general online convex optimization problem where the action \xulie\ is required to be monotone.

Other \constr s on the arm sequence motivated by practical problems are also considered in the literature. 
For example, motivated by the concern that customers are hostile to frequent price changes (see e.g. \cite{dholakia2015}), \cite{cheung2017dynamic,perakis2019dynamic} and \cite{simchi2019phase} considered pricing policies with a given budget in the number of price changes . 
Motivated by clinical trials, \cite{garivier2017thresholding} studied the best arm identification problem when the reward of the arms are monotone and the goal is to identify the arm closest to a threshold.

%% file: model.tex
{\clean Given a discrete time horizon of $n$ rounds. 
In each round $t\in [n]:=\{1,\dots,n\}$, the seller selects a price $X_t \in [p_{\min}, p_{\max}]$ and receives a (random) demand $D_t$ \sat ing $\ho{E}[D_t] = d(X_t)$ where $d:[p_{\min}, p_{\max}]\rar [0,1]$ is called the {\em demand function}.
To reduce clutter, we assume that $p_{\min}=0$ and $p_{\max}=1$.
The seller then chooses the next price based on the observations so far. 
This decision-making procedure can be formalized as a {\em policy}.

\bdefn[Markdown policy]{}
For each $t\in [n]$, let $\mathcal{F}_t=\sigma(X_1,D_1,\cdots,X_{t-1},D_{t-1})$ and define the filtration $\mathcal{F} = (\mathcal{F}_t)_{t\in [n]}$.
A {\em policy} is an $\mathcal{F}$-predictable stochastic process $X=(X_t)_{t\in [n]}$ with $X_t\in [0,1]$ almost surely (a.s.) for all $t\in [n]$. 
Furthermore, $X$ is a {\em markdown} policy if $X_{t-1} \ge X_t$  a.s. for each $t=2,\dots,n$.
\edefn

We assume that there is an unlimited supply.
Thus, when observing the demand $D_t$, the seller can fulfill all $D_t$ units and thereby receive a {\em revenue} of $R_t = D_t \cdot X_t$.
It is sometimes convenient to work with the {\em revenue \func} $r(x):= d(x) \cdot x$. 
One can easily verify that
$\ho{E}[R_t|\mathcal{F}_{t-1}] = r(X_t)$.
The seller aims to maximize the total expected revenue $R :=\sum_{t=1}^n R_t$. 

If the true demand \func\ is known, the optimal policy is to always select the revenue-maximizing price, formally given below.

\begin{assumption}[Unique optimal price]
We assume that the revenue \func s $r$ has a unique maximum $p^*(r)$.
\end{assumption}

The uniqueness assumption is adopted for the sake of simplicity of analysis and is not essential. 
It is quite common in the literature, see, e.g., Assumption 1(b) of \citealt{broder2012dynamic} and Section 2 of \citealt{den2013simultaneously}.
 

When the demand function $d$ is unknown, we measure the performance of a policy by its loss due to not knowing $d$.

\begin{definition}[Regret]
The {\it regret} of a policy $X=(X_t)$ for demand \func\ $d$ is \[\mathrm{Reg}(X, d) := n\cdot \max_{p\in [0,1]} \{p\cdot d(p)\} - \ho{E}\lb[\sum_{t=1}^n X_t \cdot d(X_t)\rb].\]
The {\it regret} of relative to a family $\mathcal{F}$ of demand functions is
\[\mathrm{Reg}(X,\mathcal{F}) := \sup_{d\in \mathcal{F}} \lb\{\mathrm{Reg}(X, d)\rb\}.\] 
\end{definition}

We first introduce a standard \assu\ (see, e.g., Lemma 3.11 in \citealt{kleinberg2003value} and Corollary 2.4 in \citealt{broder2012dynamic}) that the derivative of the revenue function vanishes at an optimal price.

\begin{assumption}[First-order optimality]\label{assu:interior_opt}
The revenue \func\ $r:[0,1]\rar [0,1]$ is differentiable and $r'(p^*(r)) = 0$.
\end{assumption}


As a standard \assu\ in the literature of MAB, in order to apply concentration bounds, we assume that the random rewards have light tail. 

\begin{definition}[Subgaussian Random Variable]
The {\it subgaussian norm} of a \rv\ $X$ is $\|X\|_{\rm sg}:=\inf \{c> 0: \ho{E} [e^{X^2/c^2}] \leq 2\}.$
We say $X$ is {\it subgaussian} if $\|X\|_{\rm sg}<\infty$.
\end{definition}

\begin{assumption}[Subgaussian noise]\label{assu:subg}
There exists a constant $c_{sg}>0$ such that the \rand\ demand $D$ at any price \sats\ $\|D\|_{\rm sg}\le c_{sg}$.
\end{assumption}

We clarify some notation before ending this section.
Throughout, we will prioritize the use of capital letters for \rv s and sets.
We will use boldface font to denote vectors. 
The notation $\tilde O$ means that we ignore the $O(\log n)$ terms.
}

%% file: crossing.tex
{\clean We characterize the complexity of a family of demand models by introducing the {\em crossing number}.
The formal definition of the {\em crossing number} may be slightly technical. 
We first illustrate the ideas using  linear demand functions, and  define the crossing number informally at the end of Section \ref{subsec:lin}. 
Readers may choose to skip the remainder of this section during initial reading without affecting their understanding of the rest of the paper.

\subsection{An illustrative example: linear demand}\label{subsec:lin}
Given $\theta_1,\theta_2>0$, consider the demand function \[d(x;\theta) = \frac 12 + \theta_1 -\theta_2 x \quad \text{for}\quad x\in \lb[\frac 12, 1\rb]\]
and consider the family $\mathcal F = \lb\{d(x;\theta): \theta_1,\theta_2 \in \lb[\frac 12, 1\rb]\rb\}$.
Here is a natural markdown policy. 
Choose two {\em exploration prices} $p_1,p_2$ close to $1$, and select each of them a number of times.
Denote by $\bar D_1, \bar D_2$ the corresponding \emp\ mean demands.
Let $\hat \theta\in \real^2$ be the unique vector that satisfies $d(p_i; \hat \theta) = \bar D_i$ for $i=1,2$. 
Select the optimal price of $d(\cdot;\hat\theta)$ in the remaining rounds. 

We observe two key properties the above parameterized family. 
First, any two curves in the family intersects at most once. 
Consequently, the true demand function can be identified using only two prices, assuming that there was no noise in the realized demands.
Second, the parameterization is  {\em robust}: \Sps\ each $\bar D_i$ deviates from the mean demand $d(p_i;\theta)$ by at most $\delta$, then the estimation error in $\theta$ is $O(\delta /|p_1-p_2|)$. 
In other words, the estimation error is (at most) linear in $\delta$ and the {\em reciprocal} of the ``dispersion'' of the exploration prices. 
As we will formally define shortly, any parameterized family with the above two properties is said to have {\em crossing number} $1$.

In broader terms, informally, a parameterized family has a crossing number $k$, if (i) any two curves intersect at most $k$ times and (ii) there exists a robust parametrization.
Specifically, (ii) means that for any set of $k$ prices spaced at a distance of $h$ from each other, an $O(\delta)$ error in the estimated demands leads to an estimation error in the parameters of $O(\delta h^{-k})$.
Under this definition, the family of degree $k$ polynomials has a crossing number $k$, under mild \assu s. 

We formalize the notion of the crossing number by formally defining two concepts: the {\it identifiability} of a family (in Section \ref{subsec:iden}) and the {\it robustness} of a \pmtn\ (in Section \ref{subsec:robust_pmtn}).

\subsection{Identifiability of a Family}\label{subsec:iden}

Consider linear demand.
\Sps\ $c\in (0,1)$ is an \unk\ constant and the demand \func\ is $d(p;c) = 1+ cp$ for $p\in [0,1]$. 
Then, we can learn the demand function by selecting just one price. 
On the contrary, consider $d(p;a,b) = a-bp$ where both $a,b$ are \unk. 
Then, we need to select (at least) two distinct prices.
Consequently, we face the following dilemma, caused by the monotonicity constraint.
\Sps\ these two exploration prices $p < p'$ are close by. 
Then, a substantial sample size is needed for reliable estimation, which leads to a high regret.
\OTOH, if $p,p'$ are far apart, then we face an {\em overshooting risk}: If the optimal price $p^*$ is close to $p'$, a high regret is incurred in the remaining rounds, since we can not increase the price.


The above discussion suggests that the complexity of a family depends on the number of prices required to identify a demand function. 

\bdefn[Profile Mapping]
Consider a set $\mathcal{F}$ of real-valued \func s defined on  $[0,1]$. 
For any ${\bf p} = (p_0,p_1,\dots,p_k)\in [0,1]^{k+1}$, the {\it profile mapping} is defined as  
\begin{align*}
\Phi_{\bf p}: \mathcal{F} &\rar \real^{k+1},\\
d &\mapsto \left(d(p_0),d(p_1),\dots,d(p_k)\right).
\end{align*}
\edefn

We refer to $\Phi_{\bf p}(d)$ as the {\it $\bf p$-profile} of a \func\ $d$.
A family of \func s is {\em $k$-identifiable} if any two functions have distinct ${\bf p}$-profile for any ${\bf p} \in [0,1]^{k+1}$.
Geometrically, this means that the graphs of any two \func s (i.e., ``curves'') intersect at most $k$ times. 

\bdefn[Identifiability]
A family $\mathcal{F}$ of \func s defined on $[0,1]$ is {\it $k$-identifiable}, if for any {\it distinct} prices ${\bf p}= (p_0,p_1,\dots,p_k)\in [0,1]^{k+1}$, the profile mapping $\Phi_{\bf p}$ is injective, i.e., for any distinct $f,\tilde f \in \cal F$, we have
\[\Phi_{\bf p}(f)\neq \Phi_{\bf p}(\tilde f).\]
\edefn

For example, the family of all degree $k$ polynomials is $k$-identifiable. 
We will soon use the following fact: If a family is $k$-identifiable, then for any distinct $p_0,p_1,\dots,p_k$, the profile mapping admits an inverse $\Phi_{\bf p}^{-1}:\mathcal{R}_p \rar \mathcal{F}$ where $\mathcal{R}_p$ is the range of the mapping $\Phi_{\bf p}$.

\subsection{Robust Parametrization}\label{subsec:robust_pmtn} 
Apart from the identifiablity, we also require the family to admit a well-behaved representation, called a {\em robust \pmtn}. 
To formalize this concept, we first define a {\it \pmtn}.

\bdefn[Parametrization] Let $k\ge 1$ be an integer and $\Theta\subseteq \real^k$.
An {\it order-$k$ \pmtn} for a family $\mathcal{F}$ of \func s is any one-to-one mapping from a set $\Theta$ to $\mathcal{F}$.
Moreover, each $\theta\in \Theta$ is called a {\it \pmt}.
\edefn

We use $d(p;\theta)$ to denote the \func\ in $\cal F$ that \pmt\ $\theta$ corresponds to.
For example, for $d(p;\theta_1,\theta_2) =\theta_1 - \theta_2 p$ is a \pmtn\ of the family of linear \func s.
As a standard \assu\ (see, e.g., page 3 of \cite{broder2012dynamic}), we assume that $\Theta$ is compact. 
\begin{assumption}[Compact Domain]\label{assu:compact}
The domain $\Theta$ of the \pmtn\ is compact.
\end{assumption}

This \assu\ leads to several favorable properties.
For example, the \func s in $\mathcal{F}$ are bounded, so we lose no generality by scaling the range of those \func s to $[0,1]$.

\begin{assumption}[Smoothness]\label{assu:smooth}
The function $d:[0,1]\times \Theta \rar \real$ is twice continuously differentiable.
Consequently, there exist constants $c^{(j)}>0$, $j=0,1,2$, such that $|d^{(j)}(p,\theta)|\leq c^{(j)}$ for any $(p,\theta)\in [0,1]\times \Theta$.
\end{assumption}

By abuse of notation, we view the optimal price mapping $p^*$ as defined on $\Theta$ rather than on $\cal F$. 
For example, for $d(p;\theta_1,\theta_2) = \theta_1 - \theta_2 p$, one can verify that $p^* = \frac{\theta_1}{2\theta_2}$.
The next \assu\ allows us to propagate the estimation error in $\theta$ to that of $p^*(\theta)$.

\begin{assumption}[Lipschitz Optimal Price Mapping]\label{assu:well-behaved}
The optimal price mapping $p^*:\Theta\rar [0,1]$ is
$c^*$-Lipschitz for some constant $c^*>0$.
\end{assumption}

This assumption has appeared in the previous literature on parametric demand learning; see, e.g., Assumption 1(c) of \citealt{broder2012dynamic}. 
Moreover, it is satisfied by many basic demand functions such as linear, exponential and logit functions.
For \ins, let $d(p;c) = 1-cp$ where $p,c\in [0,1]$, then $p^*(c) = \min\{\frac 1{2c},1\}$, which is $1$-Lipschitz.

The final ingredient for robust \pmtn\ is motivated by a nice property of the {\it natural \pmtn} $D(p;\theta) = \sum_{j=0}^k \theta_j p^j$ for \dxs s.
Consider any distinct $p_0,p_1,\dots,p_k$, and real numbers $y_0,y_1,\dots,y_d$ representing the mean reward at each $p_i$. 
We can uniquely determine a degree-$k$ \dxs\ by solving the linear equation 
\[V_{\bf p} {\bf \theta} = y, \quad\text{where}\quad {\bf \theta} = (\theta_0,\theta_1,\dots,\theta_k)^T, \quad {\bf y} = (y_0,y_1,\dots,y_k)^T\] and 
\[ V_{\bf p}:=V(p_0,\dots,p_k) = 
\begin{bmatrix}
&1 &p_0  &p_0^2& \cdots &p_0^k\\
 &1 & p_1 & p_1^2& \cdots &p_1^k\\
 & & &\vdots & &\\
&1 & p_k & p_k^2& \cdots &p_k^k\\
\end{bmatrix}\] is the {\it Vandermonde} \mtx. 
One can easily verify that $V_{\bf p}$ is invertible if and only if $p_i$'s are distinct, in which case we have \[\theta = V_{\bf p}^{-1} {\bf y}.\]
Next we consider the impact of a perturbation on $\bf y$, in terms of the following {\it dispersion} \pmt.

\bdefn[Dispersion of exploration prices]
For any ${\bf p}=(p_0,\dots,p_k)\in [0,1]^{k+1}$, we define the {\em dispersion} $h({\bf p}) := \min_{i\neq j} |p_i - p_j|.$
\edefn

To motivate the general definition of robustness, we first consider a result specific to polynomials.
Recall that $\mathcal{R}_{\bf p}$ is the range of the profile mapping $\bf \Phi_p$.
\begin{proposition}[Robust  Parametrization for Polynomials]
\label{prop:jan3}
There are constants $c_1,c_2>0$ such that for any ${\bf p}\in [0,1]^{k+1}$ with $0< h({\bf p})\leq c_1$, and ${\bf y,\hat y}\in \mathcal{R}_{\bf p}$ with $\|{\bf y}-{\bf \hat y}\|_1\le c_1$,  it holds that
\begin{align}\label{eqn:jan4}
\|V^{-1}_{\bf p} {\bf y} -V^{-1}_{\bf p}{\bf \hat y}\|_1 \leq c_2 \cdot \|{\bf y}-{\bf \hat y}\|_1 \cdot h({\bf p})^{-k}.
\end{align}
\end{proposition}

More concretely, let $d(p;\theta)$ be the true demand \func, then ${\bf y}= V_{\bf p} \theta$ represents the mean demands at the prices in $\bf p$. 
\Sps\ we observe \emp\ mean demands ${\bf\hat Y}=(\hat Y_0,\hat Y_1,\dots,\hat Y_k)$ at these prices, then we have a reasonable estimation $\hat \theta = V^{-1}_{\bf p} \bf \hat Y$.
Our Proposition~\ref{prop:jan3} can then be viewed as an upper bound on the estimation error $\|\theta - \hat\theta\|_1$ 
in $h(\bf p)$ and $\|\bf y-\bf \hat Y\|_1$.


In order to achieve sublinear regret, we need to ensure that $h({\bf p}) = o(1)$ as $n\rar \infty$.
Moreover, the rate of this convergence crucially affects our regret bounds. 
Proposition~\ref{prop:jan3} establishes a nice property for polynomials, that the estimation error scales as $h^{-k}$ when  $h\rar 0^+$.
We introduce {\it robust \pmtn} by generalizing this property beyond polynomials. We say an order-$k$ \pmtn\ is {\it robust} if the bound in Proposition \ref{prop:jan3} holds.

\bdefn[Robust Parametrization]\label{def:robust}
An order-$k$ \pmtn\ $\theta: \Theta\rar \mathcal{F}$ is {\it robust}, if \\
(1) it \sats\ Assumptions \ref{assu:compact},~\ref{assu:smooth} and \ref{assu:well-behaved}, and \\
(2) there exists $c_1, c_2>0$ such that for any ${\bf p}\in [0,1]^{k+1}$ with $0< h({\bf p})\leq c_1$ and any ${\bf y,y'}\in \mathcal{R}_{\bf p}$ with $\|{\bf y}-{\bf y'}\|_1 \leq c_1$, we have
\begin{align}\label{eqn:jul31_2}
\|\Phi^{-1}_{\bf p}({\bf y})-\Phi^{-1}_{\bf p}({\bf y'})\|_1 \leq c_2 \cdot \|{\bf y}-{\bf y'}\|_1 \cdot h({\bf p})^{-k}.
\end{align}
In \parti, when $k=0$, this \ineq\ simply means that ${\bf \Phi_p^{-1}}$ is $c_2$-Lipschitz.
\edefn

\subsection{Crossing Number of a Family}
Now we are ready to define the crossing number. 
\bdefn[Crossing number]
The {\it crossing number} of a family $\mathcal{F}$ of \func s, denoted $\kappa(\mathcal{F})$, is the minimum integer $k\geq 0$ such that $\mathcal{F}$ (i) is $k$-identifiable and (ii) admits a robust order-$k$ \pmtn. 
If no finite $k$ satisfies the above, then $\kappa(\mathcal{F}):=\infty$. 
A family with a crossing number $k$ is also called {\em $k$-crossing}.
\edefn

We illustrate our definition by considering the crossing numbers of some basic families.
As one important example, our definition of the $0$-crossing family is equivalent to the {\it separable} family defined in Section 4 of \citealt{broder2012dynamic}.
Here, we provide a few concrete examples, with proofs deferred to Appendix \ref{apdx:model}.

\bprop[$0$-crossing Families]\label{prop:examples} 
The following families are $0$-crossing:\\
{\rm (1)} The  single-parameter linear  family $\mathcal{F}_1=\{d_a(x): a\in \lb[\frac 12, 1\rb]\}$ where $d_a(x) = 1-ax$ for $x\in [\frac 12,1]$,\\
{\rm (2)} The exponential family  $\mathcal{F}_2=\lb\{d_a(x): a\in \lb[\frac 12, \frac 34\rb]\rb\}$ where  $d_a(x)=e^{1-ax}$ for $x\in [\frac 12, 1]$, and\\
{\rm (3)} The logit family $\mathcal{F}_3=\lb\{D_a(x): a\in \lb[\frac 12, 1\rb]\rb\}$ where $d_a(x) = \frac{e^{1-ax}}{1+e^{1-ax}}$ for $x\in \lb[\frac12, 1\rb]$.
\eprop

\begin{proposition}[Degree-$k$ polynomial family is $k$-crossing]
\label{prop:apr6}
Define $d(x;\theta)=\sum_{j=0}^k \theta_j x^j$. \Sps\ $\mathcal{F}=\{d(x;\theta):\theta\in \Theta\}$ \sats\ Assumptions \ref{assu:compact}, \ref{assu:smooth} and \ref{assu:well-behaved}, then $\kappa(\mathcal{F})=k$.
\end{proposition}

We defer the proof to Appendix \ref{apdx:model}.
Finally, note that if a family is not $k$-identifiable for any $k$, then $\kappa(\cal F)=\infty$. 
One example is the Lipschitz family.
In fact, for any $k< \infty$ distinct prices, there exist multiple (more precisely, infinitely many) Lipschitz \func s having the same values on these $k$ prices.

\begin{proposition}[Lipschitz family is $\infty$-crossing]\label{prop:aug2}
Let $\mathcal{F}$ be the family of $1$-Lipschitz \func s on $[0,1]$, then $\kappa(\mathcal{F}) = \infty$.
\end{proposition}

\subsection{Sensitivity} 
Finally, we introduce the notion of {\em sensitivity} to refine our categorization of families. 
For intuition, consider the Taylor expansion of a revenue \func\ $r(p)$ around $p^*$:
\[r(p)=r(p^*) + 0+ \frac{1}{2!}r''(p^*)(p-p^*)^2 + \frac{1}{3!}r^{(3)}(p^*)(p-p^*)^3 +\cdots,\]
where the ``$0$'' follows from the first-order optimality \cond\ (Assumption \ref{assu:interior_opt}).
\Sps\ the first nonzero derivative is $r^{(j)}(p^*)$. Then, the higher $j$, the less {\it sensitive} the revenue is to the estimation error in $p^*$.
We capture this aspect in the following notion of {\it sensitivity}.
\bdefn[Sensitivity]
A revenue \func\ $r$ is  {\it $s$-sensitive} if it is $(s+1)$-times continuously differentiable with
$r^{(1)}\left(p^*(r)\right) = \cdots = r^{(s-1)}\left(p^*(r)\right) = 0$ and $r^{(s)}\left(p^*(r)\right) < 0$.
A family $\mathcal{F}$ of reward \func s is called {\it $s$-sensitive} if \\
(a) every $r\in \mathcal{F}$ is $s$-sensitive,\\
(b) it admits a \pmtn\ $r(p;\theta)$ satisfying Assumptions \ref{assu:compact} to \ref{assu:well-behaved},\\
(c) there is a constant $c_6>0$ such that
$r^{(s)}\left(p^*(r)\right) \leq -c_6 < 0$ for any $r\in \mathcal{F}$, and \\
(d) for each $0\leq j\leq s$, there exists a constant $c^{(j)}$ such that $|r^{(j)}(x;\theta)|\leq c^{(j)}$, $\forall x\in [0,1]$ and $\theta\in\Theta$. 
\edefn

For example, let $s\geq 3$ and $r(x;\theta) = \theta - |\frac 12 - x|^s$ for $x\in [0,1]$ and any $\theta\in\real$, then $\{r(x;\theta):\theta \in [\frac 12,1]\}$ is an $s$-sensitive family.
Note that by Taylor's Theorem, we have 
\[|r(p^* + \eps)- r(p^*)| =  O(\eps^s).\]
Consequently, if a policy undershoots or overshoots the optimal price by $\eps$, then the regret {\it per round} is $O(\eps^s)$, which is  lower than the per-round regret $O(\eps^2)$ in the basic case.
In each of the next three sections, we will first present the regret bounds for the basic case $s=2$, and then characterize how this bound improves as $s$ increases.}

%% file: results_and_outline.tex

\section{Zero-crossing Family}
{\clean \cite{broder2012dynamic} considered the an important special case of the demand learning problem under the so-called {\em well-separated} \assu\ {\bf without} the monotonicity \constr; see their Section 4. 
Loosely, this definition states that at any price, any two demand functions are distinguishable.
This definition closely resembles our definition of a $0$-crossing, with a slight and non-essential difference, wherein they assumed that the Fisher information is bounded away from $0$, while we assume that the inverse profile mapping is well-conditioned.
Despite this slight technical discrepancy, it is straightforward to verify that their analysis for well-separated families remains valid for our $0$-crossing families.

Specifically, they proposed the following {\em MLE-greedy} policy: In each round, we estimate the true \pmt\ using the maximum likelihood estimator (MLE), and select the optimal price of the estimated demand \func. 
In their Theorem 4.8, they showed that this policy has regret $O(\log n)$.
This follows from the basic fact that the mean squared error (MSE) of MLE with a sample size $t$ is $\Theta(t^{-1})$, which implies that  
the expected total regret is $\sum_{t=1}^n O(t^{-1}) = O(\log n)$.

They also showed that this bound is minimax optimal; see their Theorem 4.5. 
More precisely, they constructed a family of linear demand functions and showed that any policy suffers $\Omg(\log n)$ regret. 
This family can be easily verified to have a crossing number of $0$; see our Proposition \ref{prop:examples}.
To facilitate a more direct comparison of their results with ours, let us express the above bounds using our terminology.

\begin{theorem}[Regret without monotonicity, $k=0$]\label{thm:BR12}
For any $0$-crossing family $\mathcal{F}$, we have ${\rm Reg}(\textrm{MLE-greedy}, \mathcal{F}) = O(\log n).$
Moreover, there exists a $0$-crossing family $\mathcal{F}$ such that for any policy $X$, we have ${\rm Reg}(X, \mathcal{F}) = \Omg(\log n).$
\end{theorem}

\begin{algorithm}[h]
\caption{Cautious Myopic (CM) Policy\label{alg:cm}}
\begin{algorithmic}[1]
\State{Input: a family $\mathcal{F}$ of demand \func s and time horizon $n$.}
\State{$P_1 \lar 1$}
\Comment{Initialization}
\For{$j=1,\dots,m$}
\For{$t=t^{(j-1)}+1,\dots,t^{(j-1)}+t_j$}\Comment{Select price $P_j$ in the next $t_j$ rounds}
\State{$X_t \lar P_j$}
\State{Observe demand $D_t$}
\EndFor
\State{$\bar D_j = \frac 1{t_j}\sum_{\tau=1}^{t_j} D_{t^{(j-1)} + \tau}$}\Comment{Empirical mean demand in phase $j$}
\State{$\hat\theta_j \lar \Phi^{-1}_{P_j}(\bar D_j)$}
\Comment{Estimate \pmt}
\State{$w_j \lar 4c_2\cdot c_{\rm sg} \sqrt{\frac {\log n}{t_j}}$}\Comment{Width of the confidence interval}
\State{$\tilde P_{j+1} \lar \max\lb\{p^*(\theta): |\theta - \hat\theta_j|\leq w_j\rb\}$}
\Comment{Conservative estimation of the optimal price}
\State{$P_{j+1}\lar \min\lb\{\tilde P_{j+1}, P_j\rb\}$} \Comment{Ensure monotonicity}
\EndFor
\end{algorithmic}
\end{algorithm}

Apparently, the price \xulie\ in the MLE-Greedy policy may not be monotone - the number of price increases can be $\Omg(n)$.
We propose the following markdown policy, dubbed the {\it Cautious Myopic} (CM) policy.
In contrast to the famous principle of {\bf optimism} under uncertainty (e.g., in UCB-type policies), our policy adopts {\bf conservatism} under uncertainty.
We partition the time horizon into {\it phases} where phase $j$ consists of $t_j= \lceil 9^j \log n \rceil$ rounds. 
It can be verified that the total number of phases is $m = O(\log n - \log\log n)$.  
In each phase, the policy builds a confidence interval $I_j$ for the true \pmt\ using the observations from (only) this phase and sets the price for phase $j+1$ to be the {\bf largest} optimal price of any ``surviving'' \pmt\ $\theta\in I_j$.
We formally state this policy in \Alg~\ref{alg:cm}. 

We show that the CM policy has regret $O(\log ^2 n)$.
Denote $t^{(j)}:=\sum^{j}_{k=0} t_k$ and $t^{(0)}=0$.

\begin{theorem}[Upper bound, $k=0$]
\label{thm:ub0d}
For any $0$-crossing family $\mathcal{F}$, we have \[\mathrm{Reg}(\mathrm{CM}, \mathcal{F})= O(\log^2 n).\]
\end{theorem}

It is worth noting that this bound is asymptotically higher than the $O(\log n)$ upper bound for unconstrained  pricing, as stated in Theorem \ref{thm:BR12}.
This is because the CM policy purposely makes conservative choices of prices to avoid overshooting.
A natural question then is: Can we achieve $o(\log^2 n)$ regret by behaving less conservatively?
We answer this question negatively, and provide a {\it separation} between markdown and  unconstrained pricing for crossing number $0$.

\begin{theorem}[Lower bound, $k=0$]
\label{thm:lb0d}
There exists a $0$-crossing family $\cal F$ of demand functions such that for any policy $X=(X_t)_{t\in [n]}$, we have \[\mathrm{Reg}(X, \mathcal{F})\ge \Omg\lb(\log^2 n\rb).\]
\end{theorem}

\begin{figure}[h]
\centering
\includegraphics[width=12cm]{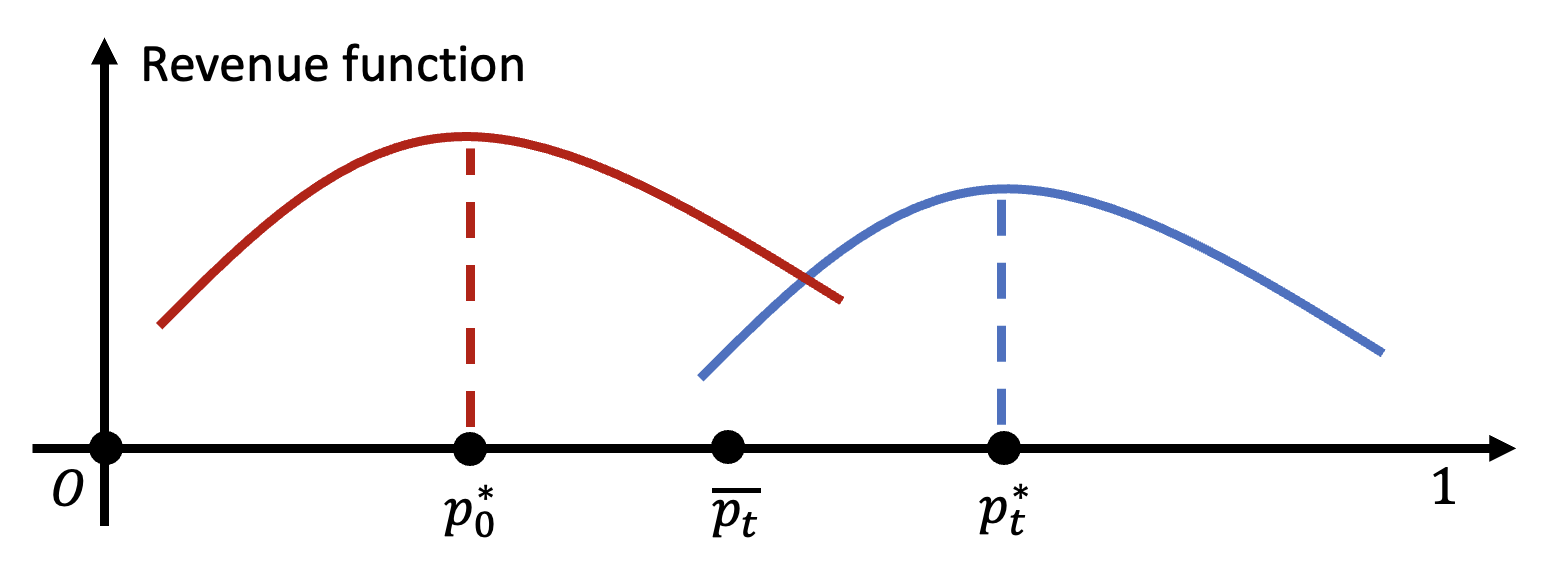}
\caption{Illustration of the lower bound proof for $0$-crossing demand family. 
The red and blue curves correspond to the revenue \func s of $d_0(\cdot)$ and $d_t(\cdot)$ for a fixed round $t$. 
Due to the monotonicity \constr, to avoid high regret for the blue curve, we need the event $X_t> \bar p_t$ to occur w.h.p. 
This in turn implies that $X_t>\bar p_t$ also occurs with $\Omg(1)$ \prb\ for the red curve, leading to a high regret in this round.}
\label{fig:6delta}
\end{figure}

We describe the proof at a high level and defer the details to Appendix \ref{apdx:lb_d=1}.
Fix a round $t= O(\sqrt n)$ and a linear demand function $d_0(x;\theta) = 1- \theta x$ whose optimal price \sats\ $p^*_0 \in (\frac 12,1)$. 
We bound the expected regret in round $t$ as follows.
Choose $\delta_t= O(\sqrt{t^{-1}\log n})$ and construct another demand \func\ $d_t(x) = 1- \theta_t x$ whose optimal price $p^*_t$ \sats\ $p^*_t \ge p^*_0+\delta_t$; see Figure \ref{fig:6delta}.

Consider the midpoint $\bar p= \bar p_t :=(p^*_t + p^*_0)/2$.
To avoid a high regret for $d_t(\cdot)$, the price $X_t$ at time $t$ must be greater than $\bar p$ with high \prb\ (w.h.p.).
In fact, due to the monotonicity \constr, once the price is lower than $\bar p$, we can not increase it back to the neighborhood of $p^*_t$, which leads to an $\Omg(\delta_t^2)$ regret in {\bf all} future rounds.
More precisely, if $X_t < \bar p_t$ occurs with \prb\ $\Omg(n^{-1/2}\log n)$, then the total regret in the remaining $n-t$ rounds is  
\[\Omg\lb(\frac{\log n}{\sqrt n}\rb)\cdot \Omg(\delta_t^2) \cdot (n-t) = \Omg\lb(\frac{\log n}{\sqrt n}\cdot \frac {\log n}t \cdot n \rb)= \Omg\lb(\log^2 n\rb),\]
where we used the assumption that $t=O(\sqrt n)$.

Therefore, to achieve low regret, the policy must exercise exceptional caution to go below $\bar p_t$.
However, this in turn leads to a high regret under $d_0$.
Using a result on the sample complexity of adaptive hypothesis testing policies due to \cite{wald1948optimum}, we can show that $X_t > \bar p_t$ occurs with \prb\ $\Omg(1)$ when $d_0$ is the true demand model.
Furthermore, when this event occurs, we incur a regret of $\Omega(\delta_t^2)$.
Note that this argument holds for any $t=1,\dots, \sqrt n$, so we can lower bound the total regret as \[\sum_{t=1}^{O(\sqrt n)} \Omg(\delta^2_t)\cdot \Omg(1) = \Omg\lb(\sum_{t=1}^{O(\sqrt n)} \frac{\log n}t\rb) =\Omg\lb(\log^2 n\rb).\]

We end this section with a natural generalization.
Recall that any family \sat ing our regularity \assu s (Assumption \ref{assu:smooth}) has sensitivity $s\ge 2$. 
Interestingly, we show that if $s>2$, the upper bound can be improved to $O(\log n)$, which matches the optimal regret bound {\bf without} the monotonicity \constr.

\begin{theorem}[Zero-crossing Upper Bound, higher sensitivity]
\label{thm:ub0d_s}
Let $s>2$ and $\mathcal{F}$ be a $0$-crossing, $s$-sensitive family of demand \func s. 
Then, \[\mathrm{Reg}(\mathrm{CM}, \mathcal{F})= O(\log n).\]
\end{theorem}
So far we have characterized the minimax regret for $0$-crossing families. 
Next, we study the finite-crossing families and analyze a {\em different} conservatism-based policy.}

\section{Finite-crossing Family}
{\clean For the crossing number $k\ge 1$, the seller needs $k+1$ distinct {\it exploration prices}, as opposed to just one for $k=0$ in the last section. 
However, this brings about additional regret, since the optimal price can lie {\it between} these exploration prices.
A policy faces the following trade-off. 
\Sps\ we select exploration prices $P_0>P_1>\dots>P_k$ sequentially, each sufficiently many times.
If these prices are spread out, the policy may learn the parameter efficiently. 
However, there is potentially incurs a high regret for overshooting, which could happen if the optimal price lies close to $P_0$.
On the other extreme, if these prices are concentrated, we need a large sample size for a good estimate, which also leads to a high regret.

\subsection{The Iterative Cautious Myopic Policy}
We handle the above dilemma by the following the {\em Iterative Cautious Myopic}  (ICM) policy, formally stated in \Alg~\ref{alg:icm}.
This policy is specified by three types of policy \pmt s: (i)  the maximum number $m$ of exploration phases, (ii) the distance $h$ between neighboring exploration prices, and (iii) the numbers $n_j$ of times we select each exploration price in  phase $j=1,\dots,m$.
We emphasize that $h$ is the same for all phases, but $n_j$ can vary between phases $j$.

\begin{figure}[h]
\centering
\includegraphics[width=11cm]{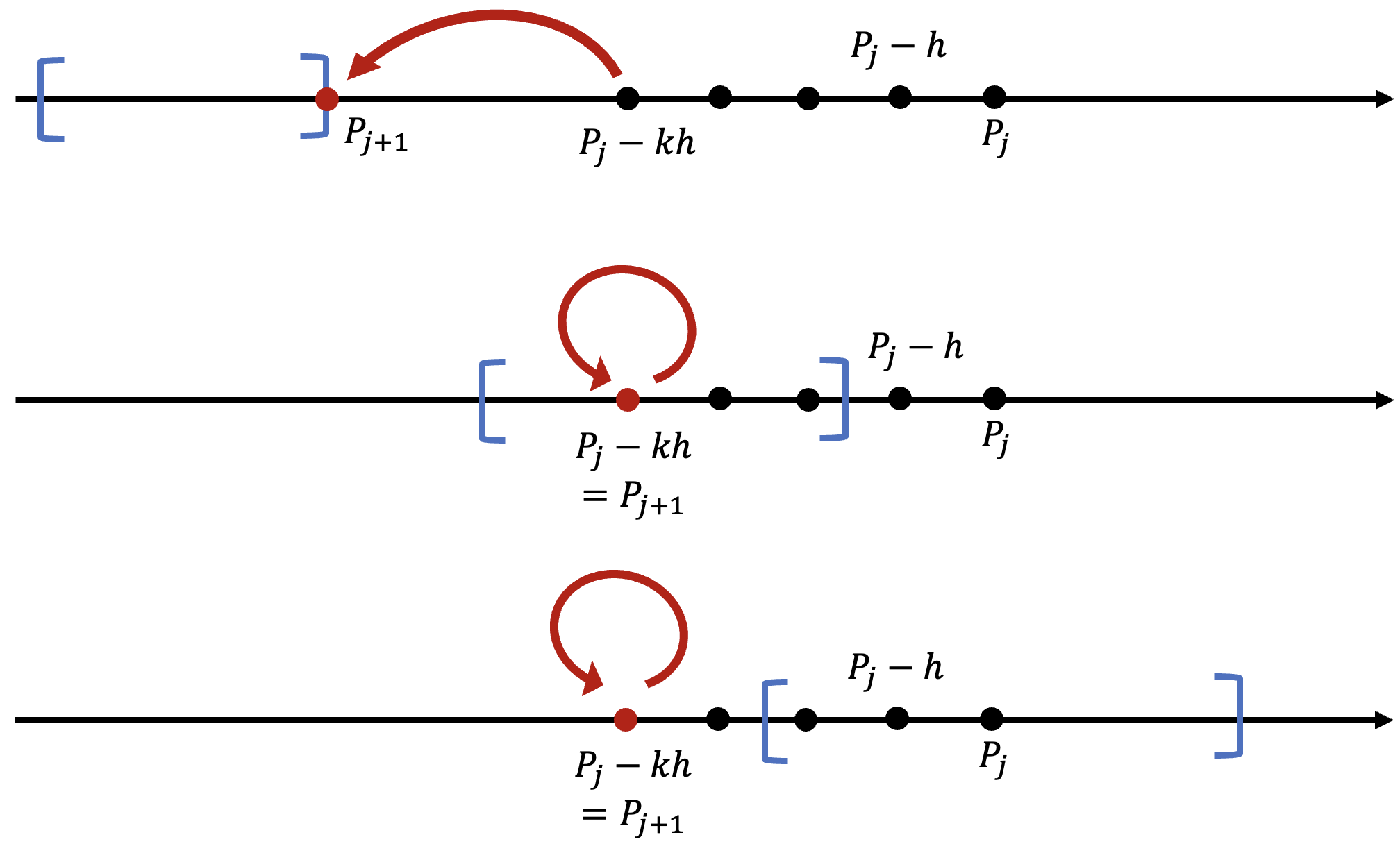}
\caption{The three subfigures correspond to the good, dangerous and bad events \resp. The blue interval represents the confidence interval $[L_j,U_j]$, obtained using observations from the exploration prices $P_j, P_j-h, \dots, P_j - kh$. The red arrow indicates the starting exploration price in the next phase, i.e., $P_{j+1}$.}
\label{fig:good_dangerous_bad}
\end{figure}

The policy starts at the price $P_0= 1$.
In phase $j\in [m]$, the policy selects prices $P_j, P_j-h,\cdots,P_j - kh$ each for $n_j$ times.
Based on the observed demands at these prices, the policy computes
a confidence interval $[L_j,U_j]$ for the optimal price.
To determine the starting price $P_{j+1}$ in the next phase, we consider the relation between the confidence interval $[L_j,U_j]$ and the current price, i.e., $P_j - kh$.
Consider the following three cases, as illustrated in Figure \ref{fig:good_dangerous_bad}.\\
1. {\bf Good event.} If $P_j - kh > U_j$, then we get closer to the optimal price by setting
$P_{j+1}\lar U_j$.\\
2. {\bf Dangerous event.} If $L_j \leq P_j - kh \leq U_j$, i.e., the current price is already inside the confidence interval. 
To avoid overshooting the optimal price, we behave conservatively by choosing $P_{j+1}$ to be the highest price allowed, i.e., $P_{j+1} \lar P_j - kh.$\\
3. {\bf Overshooting event.} If $P_j - kh< L_j$, then w.h.p. we have overshot, i.e., 
the current price is lower than $p^*$, so we immediately terminate the exploration phase and select the current price in all remaining rounds.

\begin{algorithm}[h]
\begin{algorithmic}[1]
\State{Input: \\
\quad $\mathcal{F}:$ a family of demand functions,\\
\quad $m$: number of phases,\\ 
\quad $n_1,\dots,n_m$: length of each phase,\\
\quad $n:$ length of the time horizon.}
\State{Initialization: $P_1\lar 1, L_0\lar 0, U_0\lar 1$}
\For{$j=1,2,\dots,m$}\Comment{Phase $j$}
    \For{$i =0,1,\dots,k$}\Comment{Exploration price $i$}
        \State{Select price $P_j-i  h$ for $n_j$ times in a row and observe demands $D_1,\dots,D_{n_j}$}
        \State{$\bar D_\ell\lar \frac 1{n_j} \sum_{i=1}^{n_j} D_i$\Comment{Empirical mean demand at $P_j-ih$}}
    \EndFor
    \State{$\hat \theta \lar\Phi^{-1}_{P_j,\dots,P_j-kh} \lb(\bar D_0,\dots,\bar D_k\rb)$ \Comment{Estimate the parameter}}
    \State{$w_j\lar 2h^{-k} c_2\cdot c_{\rm sg} \sqrt \frac {k\log n}{n_j}$
    \Comment{ Width of confidence interval}}
    \State{$L_j\lar \min\{p^*(\theta): \|\theta - p^*(\hat \theta)\|_2 \le w_j\}$ 
    \Comment{Lower confidence bound}}
    \State{$U_j\lar \max\{p^*(\theta): \|\theta - p^*(\hat \theta)\|_2 \leq w_j\}$}
    \Comment{Upper confidence bound}    \If{$U_j > P_j-kh$} 
    \State{$P_{j+1}\lar U_j$
    \Comment{Good event}} 
    \EndIf
    \If{$U_j \geq P_j-kh \geq L_j$}    \State{$P_{j+1} \lar P_j - kh$
    \Comment{Dangerous event}}
    \EndIf
    \If{$P_j - kh < L_j$} 
    \State{Break\Comment{Overshooting event}
    } 
    \EndIf
\EndFor
\State{Select the current price in every future round}
\Comment{Exploitation}
\caption{The Iterative Cautious Myopic (ICM) Policy
\label{alg:icm}}
\end{algorithmic}
\end{algorithm}

\subsection{Regret Analysis of the ICM Policy}
We first present a general upper bound on the regret of the ICM policy for \arb\ $h$ and $n_j$'s. 

\begin{proposition}[Upper bound, finite-crossing family]\label{prop:ub_gen_d}
Let $\mathcal{F}$ be a $k$-crossing, $s$-sensitive  family of demand functions where $s\ge 2$. 
\Sps\ $h\in (0,1)$ and $1=: n_0 <n_1<\dots <n_m$ where $n_m=o(n)$. 
Denote $\mathrm{ICM} = \mathrm{ICM}(n_1,\dots,n_m,h)$. Then, 
\begin{align}\label{eqn:121223}
\mathrm{Reg}(\mathrm{ICM},\mathcal{F})&\le c_s \left(3c^* c_{\rm sg}k \frac{\sqrt{c_5 k \log n}}{h^k}\right)^s \lb(\sum_{j=1}^{m-1}\frac{n_j}{n_{j-1}^{s/2}} + \frac n{n_m^{s/2}}\rb) + c_s \left(mkh\right)^s n + \frac {2mk}{n^2}\\
&=\widetilde O\lb(h^{-sk}\lb(\sum_{j=1}^{m-1} \frac{n_j}{n_{j-1}^{s/2}} + \frac n{n_m^{s/2}} \rb) + h^s n\rb).\notag
\end{align}
\end{proposition}

To clarify, in the above ``$\tilde O$'' notation, we view $m,k$ as small constants and suppressed them along with $\log n$ and constant terms.
This is done to highlight the dependence on the most important policy parameters, such as $(n_j)$ and $h$.

Let us sketch the proof before finding the optimal choice of $(n_j)$. 
\Sps\ the ICM policy enters some phase $j\ge 1$ before the exploration ends (i.e., before the overshooting event occurs).
Denote by ${\bf Y}\in \real^{\{0,1,\cdots,k\}}$ the true demand vector, i.e., \[{\bf Y} =\lb(d(P_0;\theta),d(P_k;\theta), \dots, d(P_k;\theta)\rb).\]
Similarly, let the vector ${\bf Y'}\in \real^{\{0,1,\dots,k\}}$ denote the \emp\ demand observed at the exploration prices. 
By Hoeffding's \ineq\ (formally stated in  Appendix \ref{apdx:ub}), w.h.p. each entry of ${\bf Y'}$ deviates from ${\bf Y}$ by $\tilde O(n_j^{-1/2})$, and hence 
\[\lb\|{\bf Y}-{\bf Y'}\rb\|_1 = \tilde O\lb(k n_j^{-1/2}\rb).\]
Denoting the estimated \pmt\ $\hat\theta$, then by 
eqn. \eqref{eqn:jan4} in the definition of robust \pmtn\ (Definition \ref{def:robust}), the estimation error is bounded as \[\|\theta - \hat \theta\|_1 = \|\Phi_{P_0,\dots, P_k}^{-1} ({\bf Y}) - \Phi_{P_0,\dots, P_k}^{-1} ({\bf Y'})\|_1 \le c_2 \cdot \|{\bf Y}-{\bf Y'}\|_1 \cdot h^{-k} = \tilde O\lb(k n_j^{-1/2} h^{-k}\rb).\] 
By Lipschitzness of the optimal price mapping $p^*(\cdot)$, the error of the estimated optimal price is also $\tilde O(k n_j^{-1/2} h^{-k})$.
Finally, by the definition of sensitivity and recalling that phase $j$ consists of $(k+1)n_j$ rounds, we can bound the regret in this phase as \[\tilde O\lb(k n_j \cdot (k h^{-k} n_{j-1}^{-1/2})^s\rb).\]
Similarly, setting $n_{m+1}:=\Theta(m)$, we can bound the regret in the exploitation phase as $\tilde O(kn\cdot (h^{-k}\cdot n_m^{-1/2})^{s})$.

Finally, we interpret the last term, $(mkh)^s n$, in eqn. \eqref{eqn:121223}.
This term corresponds to the overshooting risk, i.e., the regret caused by going lower than the optimal price $p^*$.
In fact, \sps\ the exploration phase ends due to the overshooting event in some phase $j\ge 1$, then $P_j - kh< L_j.$
By the construction of the \ci, we have $p^*\in [L_j, U_j]$ w.h.p.
Thus, \[|(P_j - kh)- p^* | \le kh + |L_j- U_j|.\]
Again, by the definition of sensitivity, the regret in exploitation phase is \begin{align}\label{eqn:121623}
\tilde O\lb((kh+ |L_j- U_j|)^s \cdot n\rb).
\end{align}
Finally, by straightforward algebraic manipulation, we can merge the term $|L_j-U_j|^s$ into the second term in eqn. \eqref{eqn:121223} and simplify  eqn. \eqref{eqn:121623} as $\tilde O((mkh)^s \cdot n)$.

We emphasize that this explanation is only for the purpose of high-level understanding. 
The formal proof involves handling numerous technical intricacies for each of the three events and is postponed to Appendix \ref{apdx:ub}.

With Proposition \ref{prop:ub_gen_d}, it is \strfwd\ to find the (\asymly) optimal choice of the policy \pmt s. 
In Appendix \ref{apdx:ub}, we show that this can be reduced to a linear program, which can be optimized by  choosing $n_j$'s and $h$ so that all terms in eqn. \eqref{eqn:121223} have the same \asym\ order.
This leads to the following regret bound.

\begin{theorem}[Upper bound for finite crossing number]\label{thm:ubgen}
\label{thm:ub_gen_s}
\Sps\ $m=O(\log n)$.
For any $s\ge 2$-sensitive, $k$-crossing family $\cal F$, there exists $n_1<\dots <n_m$ such that \[\mathrm{Reg}(\mathrm{ICM},\mathcal{F}) = \tilde O\left(n^{\rho(m,s,k)}\right) \quad \text{where} \quad \rho(m,s,k) = \frac{1+\left(1+\frac s2 +  ... +(\frac s2)^{m-1}\right)k}{\left(\frac s2\right)^m + \left(1+\frac s2 + ...
+ (\frac s2)^{m-1}\right)\cdot (k+1)}.\]
In \parti, if $s=2$, then for $h=n^{\frac m{m(k+1)+1}}$ and $n_j = n^{\frac{mk+j}{m(k+1)+1}}$, $j=1,\dots,m$, we have \[\mathrm{Reg}\lb(\mathrm{ICM},\mathcal{F}\rb) = \tilde O\left(n^{\frac k{k+1}}\right).\]
\end{theorem}


Prior to this work, the best known regret bound for non-parametric, twice continuously differentiable family is $\tilde O(n^{5/7})$; see Theorem 2 in \citealt{jia2021markdown}. 
Our Theorem \ref{thm:ubgen} improves the above bound to $\tilde O(\sqrt n)$ for $k=1$, and to $\tilde O(n^{2/3})$ for $k=2$.

\subsection{Lower Bound}
We show a lower bound that matches the upper bound up to $O(\log n)$ factors.

\begin{theorem}[Lower bound for finite $k\ge 1$]
\label{thm:lbgen}
For any $k\geq 2$, there exists a $k$-crossing family $\mathcal{F}$ of demand \func s such that for any markdown policy $X$, we have \[\mathrm{Reg}(X,\mathcal{F})= \Omg \lb(n^{\frac k{k+1}}\rb).\]
\end{theorem}

In our proof, for each $k\geq 1$ we construct a family of   decreasing, degree-$k$ polynomials -- which is also $k$-crossing -- on which any policy suffers regret $\Omg(n^{k/(k+1)})$. 

We conclude the section by comparing the above lower bound with the known results in the nonparametric setting. 
\cite{jia2021markdown} showed that there is a policy that achieves $\tilde O(n^{5/7})$ regret if the revenue function is unimodal and twice continuously differentiable. 
On the other hand, for $k\ge 3$, the lower bound in Theorem \ref{thm:lbgen} is $\Omg(n^{3/4})$, which is higher than $\tilde O(n^{5/7})$. 
But this is {\bf not} a contradiction, since for $k\ge 3$, the revenue \func s in our lower bound proof may not be unimodal.
}

%% file: apdx_prelim.tex
\subsection{Proof of Propositions~\ref{prop:jan3} and \ref{prop:apr6}}
{\clean  The following definition of the \mtx\ norm can be found in eqn (1.4) of  \citealt{gautschi1962inverses}.
It can be alternatively viewed as the operator norm of the corresponding linear mapping  under the $L_1$-norm.

\bdefn[Matrix norm]
For any $A\in \real^{m\times m}$, define 
\[\|A\|_{\rm op} := \max_{i\in [n]} \sum_{j\in [m]} |a_{ij}| = \sup_{\|v\|_1\le 1} \|Av\|_1.\]
\edefn

By this definition, for any vector $w$, we have $\|Av\|_1 \le \|A\|_{\rm op} \cdot \|v\|_1$.
Sometimes we need to consider other $L_p$ norms.
The following basic fact will be useful. 

\begin{lemma}[$L_p$ norm do not differ by much]\label{lem:norm}
For any $v\in \real^m$ and $p$ with $1\le p\le q\le \infty$, we have \[\|v\|_q \le \|v\| p \le m\cdot \|v\|_q\]
\end{lemma}

The inverse Vandermonde \mtx\ (IMV) admits a (somewhat complicated) closed-form expression due to \cite{gautschi1962inverses}, which leads to the following upper bound on its \mtx\ norm.
\begin{theorem}[Matrix norm of the IMV]
\label{thm:Gautschi62}
Let $V=V(x_1,\dots,x_m)$ be the  Vandermonde \mtx\  of distinct real numbers $x_1,\dots,x_m$.
Then, \[\|V^{-1}\|_{\rm op} \leq \max_{i\in [m]} \prod_{ j\neq i} \frac{1+|x_j|}{|x_j-x_i|}.\]
\end{theorem}

Let us apply this result to our problem.
Recall that for any distinct prices $(p_0,p_1,\dots,p_k)=:{\bf p}$, we defined $h({\bf p}) = \min_{i\neq j}\{|p_i-p_j|\}$.
We can bound $\|V^{-1}_{\bf p}\|_{\rm op}$ in terms of $h(\bf p)$ using Theorem \ref{thm:Gautschi62}.

\begin{corollary}[Bounding the norm of the IVM using $h({\bf p})$]\label{coro:aug1}
For any ${\bf p}=(p_0,p_1,\dots,p_k)\in [0,1]^{k+1}$ with distinct entries, it holds that
\[\|V^{-1}_{\bf p}\|_{\rm op} \leq \frac{2^k}{h({\bf p})^k}.\]
\end{corollary}
\proof{Proof.}
Fix any $i\in [m]$. 
Since $0\leq p_j\leq 1$ for any $j\in [n]$, we have $1+|p_j|\leq 2$. Therefore,
\[\prod_{j=1, j\neq i} \frac{1+|x_j|}{|x_j-x_i|} \leq \frac{2^k}{h({\bf p})^k}.\eqno \halmos\]

We now use the above to show that the family of degree-$k$ polynomials have crossing number $k$. 
Recall that $\mathcal{R}(\cdot)$ denotes the range of a mapping.

\noindent{\bf Proof of Proposition~\ref{prop:jan3} and ~\ref{prop:apr6}.}
Fix any ${\bf p}=(p_0,p_1,\cdots,p_k)\in [0,1]^{k+1}$ with distinct entries. 
Then, for any $y,y'\in \mathcal{R}(V_{\bf p})$,   we have
\[\| V_{\bf p}^{-1}y' - V_{\bf p}^{-1}y\|_1
\leq \|V_{\bf p}^{-1}\|_{\rm op} \cdot \|y'-y\|_1 \leq \frac{2^k}{h({\bf p})^k}\cdot \|y'-y\|_1,\]
where the first \ineq\ follows from Lemma~\ref{lem:norm} and the second follows from Corollary~\ref{coro:aug1}.
Proposition \ref{prop:jan3} then follows by selecting the constant $c_2$ to be $2^k$.
So far we have shown that the natural parametrization of \dxs\ is robust, and therefore $\kappa(\mathcal{F}) \le k$.
Moreover, for any $\ell < k$, the family $F$ is not $\ell$-identifiable, since there exists a degree-$k$ \dxs\ with $\ell$ real roots.
\IFT\ $\kappa(\mathcal{F}) \ge k$, and therefore we obtain Proposition \ref{prop:apr6}. \hfill
\halmos
}

\subsection{Proof of Proposition \ref{prop:examples}(a)}
We first observe that $\mathcal{F}_1$ is $0$-identifiable, i.e., for any fixed $p\in [\frac 12,1]$, the function $d(p;a)$ is injective in $a$. 
In fact, for any $a,a'\in [\frac 12,1]$, $d(p;a) = d(p;a')$ is equivalent to $p\cdot (a-a')=0$, which implies $a=a'$ since $x\neq 0$.

Next, we verify that the parametrization $a\mapsto d(\cdot;a)$ is robust. This involves verifying the following conditions, as required in Definition~\ref{def:robust}.\\
(1) {\bf Lipschitz optimal price mapping.} For any $a\in [\frac 12, 1]$, the revenue \func\ $r(p;a) = a (1-ap)$ has a unique optimal price $p^*(a)=\frac 1{2a}$, which is $2$-Lipschitz.\\
(2) {\bf Robust under perturbation.} For any fixed price $p\in [\frac 12, 1]$, by the definition of profile mapping $\Phi$, we have $\Phi_p(a) = 1-ap$. 
For any $y\in \mathcal{R}(\Phi_p)$, 
where we recall that $\mathcal{R}(\cdot)$ denotes the range of a mapping, we have $\Phi_p^{-1}(y)=(1-y)/p.$
Thus, for any $y,y'\in \mathcal{R}(\Phi_p)$, we have 
\[\lb|\Phi_p^{-1}(y)-\Phi_p^{-1}(y')\rb| = \lb|\frac{1-y}p - \frac{1-y'}p\rb| = \frac{|y-y'|}p \leq 2\cdot |y-y'|.\]
Therefore, the parametrization $a\mapsto D(\cdot;a)$ 
is robust, and therefore $\mathcal{F}_1$ has a crossing number $0$.\hfill \halmos

\subsection{Proof of Proposition \ref{prop:examples}(b)}
Write $d(x;a) = e^{1-ax}$.
The following will be useful for (b) and (c), whose proof follows directly from the definition of identifiability.

\begin{lemma}[Monotone link function preserves identifiablity]
\label{lem:strict}
Let $g:\real\rar \real$ be a  strictly monotone function and $\{d(x;\theta):\theta\in \Theta\}$ be a $0$-identifiable family. Then, $\{ g(d(x;\theta)): \theta\in\Theta\}$ is also $0$-identifiable.
\end{lemma}

Since $\mathcal{F}_1$ is $0$-identifiable, and $g(z) = e^z$ is strictly increasing, by Lemma~\ref{lem:strict}, $\mathcal{F}_2$ is also $0$-identifiable.
It remains to show that $a\mapsto d(\cdot; a)$ is robust for $\mathcal{F}_2$, which we verify as follows. \\
(1) {\bf Lipschitz optimal price mapping.} For any $a$, the \corres\ reward \func\ is $r(p;a) = a\cdot e^{1-ap}$.
A price $p$ is then a local optimum if and only if the derivative of $r'(p;a)=(1-ap)e^{1-ap}=0$, i.e., $x=\frac 1{2a}$, which has just been verified to be Lipschitz in the proof for (a).\\
(2) {\bf Robust under perturbation.} For any fixed price $p\in [\frac 12, 1]$, we have $\Phi_p(a) = e^{1-ap}$, and therefore for any $y\in \mathcal{R}(\Phi_p)$, we have $\Phi_p^{-1}(y)= \frac{1-\ln y}p$.
Thus, for any $y,y'\in \mathcal{R}(\Phi_p)$, 
\begin{align}\label{eqn:jul30}
\lb|\Phi_p^{-1}(y) - \Phi_p^{-1}(y')\rb| = \lb|\frac{1-\ln y}p - \frac{1-\ln y'}p\rb| = \frac{|\ln y - \ln y'|}p.
\end{align}
Finally, to bound the above, note that $e^{\frac 14} \leq e^{1-ap} \leq e^{\frac 34}$ for any $p\in \lb[\frac 12, 1\rb]$ and $a\in \lb[\frac 12, \frac 34\rb]$, i.e., \[\mathcal{R}(\Phi_p)\subseteq \lb[e^{\frac 14},e^{\frac 34}\rb].\]
Thus, for any $p\in [\frac 12,1]$,
\[\eqref{eqn:jul30} \leq  \frac{\max_{t\in \mathcal{R}(\Phi_p)}\ln' t}p \cdot \lb|y-y'\rb| = 2e^{-\frac 14}\cdot \lb|y-y'\rb|.\]
Therefore the order-$0$ parametrization $a\mapsto d(\cdot;a)$ is robust, and hence $\mathcal{F}_2$ is $0$-crossing.\hfill\halmos

\subsection{Proof of Proposition \ref{prop:examples}(c)}
The $0$-identifiability again follows from Lemma \ref{lem:strict}. 
In fact, one can easily verify that the function $g(t) = \frac{e^t}{1+e^t}$ is strictly monotone, and since $\mathcal{F}_1$ is $0$-identifiable, so is $\mathcal{F}_2$.
It remains to show that $a\mapsto d(\cdot; a)$ is robust, which can be done by verifying the following.\\
(1) {\bf Lipschitz Optimal Price Mapping.} 
Note that \[r'_a(p) = \frac{e\lb(e^{ap}\lb(1-ap\rb)+e\rb)}{\lb(e^p + e \rb)^2},\] 
so $r'_a(p)=0$ if and only if $e^{ap}(1-ap)+e=0$, i.e., 
$-(1-ap)=e^{1-ap}.$
Letting $t=1-ap$, the above becomes 
\begin{align}\label{eqn:jul29}
-t = e^t.
\end{align}
Note that RHS is increasing in $t$ whereas LHS is decreasing and surjective, so \eqref{eqn:jul29} admits a unique \sln, say, $t_0$.
Note that $p = \frac{1-t}{a}$, so $p^*(a) = \frac{1-t_0}a$.
Thus, the $p^*$ function is $2$-Lipschitz since $a\ge\frac 12$.\\
(2) {\bf Robust under perturbation.}  
For any fixed $p\in [\frac 12, 1]$, we have \[\Phi_p(a) = d(x;a) = \frac{e^{1-ap}}{1+ e^{1-ap}}.\]
By calculation, for any $y\in \mathcal{R}(\Phi_p)$ we have 
\[\Phi_p^{-1}(y) = \frac 1p \lb(1+\ln \frac {1-y}y\rb),\]
and hence for any $y,y'\in \mathcal{R}(\Phi_p)$, 
\begin{align}\label{eqn:jul31}
\lb|\Phi_p^{-1}(y)-\Phi_p^{-1}(y')\rb| &= \frac 1p\cdot \lb|\lb(1+\ln \frac {1-y}y\rb)-\lb(1+\ln \frac {1-y'}{y'}\rb)\rb|\notag\\
&= \frac 1p \lb|\ln y - \ln y' + \ln (1-y)- \ln (1-y')\rb|\notag\\
&\le \frac 1p \lb(\max_{s\in\mathcal{R}(\Phi_p)} \frac 1s \cdot |y-y'| + \max_{t\in \mathcal{R}(\Phi_p)}\frac 1{1-t}\cdot |y-y'|\rb)\notag\\
&\leq \frac{|y-y'|}p \lb(\max_{s\in\mathcal{R}(\Phi_p)} \frac 1s + \max_{t\in \mathcal{R}(\Phi_p)} \frac 1{1-t}\rb),
\end{align}
where the first \ineq\ follows from the Lagrange mean value Theorem.
To conclude, note that for any $a\in [\frac 12,1]$ and $p\in [\frac 12,1]$, we have \[\frac 12\leq \Phi_p(a)\le \frac{e^{\frac 34}}{1+e^{\frac 34}}< \frac 7{10},\]  so $\mathcal{R}(\Phi_p) \subset \lb[\frac 12, \frac 7{10}\rb]$.
Therefore,  
\[\eqref{eqn:jul31}\le 2\cdot |y-y'|\cdot \lb(2+\frac{10}3\rb) = \frac{32}3\cdot |y-y'|.\eqno\halmos\]

\subsection{Proof of Proposition~\ref{prop:aug2}}
For any given $d$, consider $f(x) = \frac{\sin(d\pi x)}{d\pi}$ and $g(x) = \frac {\sin(2d\pi x)}{2d\pi}$ and ${\bf p} = (p_i) = \lb(0,\frac 1d,\dots,\frac dd\rb)$. 
Observe that both $f$ and $g$ are $1$-Lipschitz, and moreover, $f(p_i) = g(p_i) = 0$ for any $i$.  
Therefore, $\Phi_{\bf p}$ is not injective, and therefore $\mathcal F$ is not $d$-identifiable. \hfill\halmos


%% file: ub.tex
{\clean We first state some standard results  that will be useful for our analysis.

\subsection{Preliminaries for the Upper Bounds}
Most of our upper bounds rely on the following standard concentration bound for subgaussian \rv s; see, e.g., Chapter 2 of \citealt{wainwright2019high}.
Recall from Section \ref{sec:formulation} that $\|\cdot \|_{\psi_2}$ is the subgaussian norm.

\begin{theorem}[Hoeffding's \ineq]\label{thm:hoeffding}
\Sps\ $X_1,\dots,X_n$ are \indep\ subgaussian \rv s, then for any $\delta >0$, we have
\[\ho{P}\left[\sum_{i=1}^n \left(X_i - \ho{E}X_i\right) \geq \delta\right]\leq \exp\left(-\frac{\delta^2}{2\sum_{i=1}^n \|X_i\|^2_{\psi_2}}\right).\]
\end{theorem}

We will also use a folklore result from Calculus.
\begin{theorem}[Taylor's Theorem with Lagrange Remainder]\label{thm:lag}
Let $f:\real \rar \real$ be $(m + 1)$ times differentiable on an open interval $(a,b)$.
Then for any $x,x'\in (a,b)$, there exists some $\xi$ with  $(x-\xi)\cdot (x'-\xi)\leq 0$ such that 
\[f(x') = f(x) + \frac 1{1!} f'(x) (x'-x)+ ...+\frac 1{m!}f^{(m)}(x)(x'-x)^s + \frac 1{(m+1)!}f^{(m+1)}(\xi )(x'-x)^{s+1}.\]
\end{theorem}
Theorem~\ref{thm:lag} implies a key property for any $s$-sensitive reward \func s. 
\Sps\ the revenue function $r$ is $s$-sensitive, then for any $\eps>0$, we have
\[r(p^* + \eps) = r(p^*) +  \frac{r^{(s)}(\xi)}{s!}\eps^s,\]
where $\xi \in [p^*, p^*+\eps]$.
Since $\Theta$ is compact, there exists some constant $c_s>0$ such that $|r^{(s)}(x, \theta)|\leq c_s$ for any $x\in [0,1]$ and $\theta \in \Theta$. 
Thus, \[|r(p^* + \eps)- r(p^*)|\leq \frac {c_s}{s!}|\eps|^s.\]
Consequently, if a policy overshoots or undershoots the optimal price by $\eps$, the regret {\it per round} is only $O(\eps^s)$. 
In \parti, if we only assume differentiability, then $s=2$ and the above per-round regret is $O(\eps^2)$.

\subsection{Zero-crossing Family}

Our analysis relies on a high-\prb\ error bound. 
The following lemma can be obtained as a direct consequence of Hoeffding's \ineq\ (Theorem~\ref{thm:hoeffding}).

\begin{lemma}[Clean event is likely]\label{lem:ci_0dim}
Let $Z_1,\dots,Z_m$ be i.i.d. \rv s following a \distr\ $D$ with subgaussian norm $c$. 
Let $\cal B$ be the event that
$\lb|\ho{E} [D] - \frac 1m \sum_{j=1} Z_j \rb| \le 2c \sqrt{\frac{\log n}{m}}$, then
$\ho{P}[\overline{\cal B}]\le n^{-2}$.
\end{lemma}
\proof{Proof.}
By Theorem~\ref{thm:hoeffding}, we have
\[ \ho{P}\left[\lb|\ho{E}[D] - \frac1m \sum_{j=1}^m Z_j \rb|
> 2c\sqrt{\frac{ \log n}{m}}\right]\leq \exp\left(-\frac{(2c\sqrt{ t_j \log n})^2}{2t_j \cdot c^2}\right) = n^{-2}.\eqno\halmos\]

We next apply the above to our problem by defining good events.
Recall that $c_{\rm sg}$ is the upper bound on the subgaussian norm of the demand \distr s at any price, as formalized in Assumption \ref{assu:subg}, and $\bar D_j$ is the \emp\ mean demand in phase $j=1, \dots,m = O(\log n - \log \log n)$.

\bdefn[Good and bad events]
For every $j\in [K$], let $\mathcal{E}_j$ be the event that $\big |d(P_j ;\theta^*) - \bar D_j\big | \leq 2c_{\rm sg} \sqrt{\frac{\log n}{t_j}}$. We call $\mathcal{E} = \bigcap_{j=1}^m \mathcal{E}_j$ be the {\it good event} and its complement $\mathcal{E}^c$ the {\it bad event}.
\edefn

As a standard step in regret analysis, we first show that  the bad event occurs with low \prb.

\begin{lemma}[Bad event is unlikely] It holds that 
$\ho{P}[\mathcal{E}^c]\geq 1-n^{-1}.$
\end{lemma}
\proof{Proof.}
Fix any $j\in [m]$.
Since $\{D_j: j=t^{(j-1)}+1,..., t^{(j)}\}$ is an i.i.d. sample from a subgaussian \distr\ with mean $d(P_j;\theta^*)$ and subgaussian norm at most $c_{\rm sg}$, by Lemma~\ref{lem:ci_0dim} we have $\ho{P}[\overline{\mathcal{E}_j}]\leq n^{-2}.$
By the union bound, we have
\[\ho{P}[\mathcal{E}^c] \leq n^{-2} \cdot \log n \leq n^{-1}.\eqno\halmos\]

This above implies that by conditioning on $\mathcal{E}$, we only lose only an $O(1)$ term in the regret, since the mean demand and revenue is bounded in $[0,1]$.

We next bound the estimation error in terms of observations.  
Unless stated \ow, we denote $\|\cdot\| = \|\cdot\|_\infty$.
Recall that $\Phi_{p}$ is the profile mapping, which maps a demand function $d(\cdot)$ to its value $d(p)$ at price $p\in [0,1]$.

\begin{lemma}[Bounding the estimation error]\label{lem:aug1}
\Sps\ the event $\cal E$ occurs. Then, for each $j\in [m]$,
we have 
\[\|\wh \theta_{j} - \theta^*\| \leq 2c_2 \cdot c_{\rm sg} \sqrt{\frac{\log n}{t_j}}.\]
\end{lemma}
\proof{Proof.}
By the definition robust \pmtn, we have 
\begin{align*}
\|\wh \theta_{j} - \theta^*\| = \|\Phi_{P_j}^{-1}(\bar D_j) - \Phi_{P_j}^{-1}\left(\Phi_{P_j}\lb(\theta^*\rb)\right)\| 
\leq c_2 \cdot \|\bar D_j - \Phi_{P_j}(\theta^*)\| \leq 2c_2 \cdot c_{\rm sg} \sqrt{\frac{\log n}{t_j}},
\end{align*}
where the last \ineq\ follows since $\Phi_{P_j}(\theta^*)=d(P_j,\theta^*)$ and the \assu\ that $\cal E$ occurs.
\hfill \halmos

We next show that the prices are guaranteed to be non-increasing \cond al on the good event.
To this aim, we introduce the following confidence intervals on the \pmt\ space,  centered at the estimated \pmt\ $\hat\theta_j$.
\bdefn[Confidence Interval for $\theta^*$]\label{def:CItheta}
Let $w_j = 4c_2 \cdot c_{\rm sg} \sqrt{\frac{\log n}{t_j}}$. 
Define $L_j = \wh \theta_j - w_j$, $R_j=  \wh \theta_j + w_j$. The {\em \confi\ interval} is in phase $j$ is $I_j:= [L_j, R_j]$. 
\edefn 

We clarify that the notations $L_j,R_j$ are capitalized since they are random. 
to show monotonicity, we first show that $I_j$'s form a nested \xulie. 
To this goal, we not only need to ensure that $\theta^*\in I_j$, but also that $\theta^*$ lies far from the boundary of the \ci, more precisely, it lies between the first and third quantiles of the interval.  

\bdefn[Far from the boundary]
Given a finite interval $[a,b]$ and $x\in [a,b]$, a point $x$ is {\it far from the boundary} (FFB) in $[a,b]$ if $\frac 34 a +\frac 14 b \leq x \leq \frac 14 a +\frac 34 b$.
\edefn

Note that in Definition \ref{def:CItheta}, we intentionally selected the radius to be {\it twice} as large as (rather than equal to) the bound given in Lemma~\ref{lem:aug1}, so that $\theta^*$ is FFB in $I_j$, assuming that the good event occurs.
This fact plays a crucial role in the following analysis of monotonicity.
We first observe that it suffices to show that $I_j$'s form a nested \xulie.

\begin{lemma}[Nested sequence of intervals]\label{lem:nested}
\Sps\ the good event occurs.
Then, for each $j\in [m]$, we have $I_{j+1}\subseteq I_j$.
\end{lemma}
\proof{Proof.}
We will show that $R_{j+1}\leq R_j$; the proof for $L_{j+1} \ge L_j$ is identical.
Recall that $\theta^*$ is FFB in $I_j$, so
$R_j\geq \theta^* + \frac12 w_j.$
Similarly, since $\theta^*$ is also FFB in $I_{j+1}$, we have
$R_{j+1}\leq \frac32 w_{j+1} +\theta^*.$
Combining, we have
\begin{align}\label{eqn:aug1}
    R_{j+1} \leq \frac32 w_{j+1} + R_j -\frac12 w_j.
\end{align}
Finally, we claim that $w_{j+1} \leq \frac 13 w_j$. 
This is \strfwd\ from the choice of $w_j$. 
In fact, recall that $w_j = 4c_2 \cdot c_{\rm sg} \sqrt{\frac{\log n}{t_j}}$ and $t_j = 9^j\log n$, so \[\frac {w_{j+1}}{w_j} = \sqrt{\frac{t_j}{t_{j+1}}}= \sqrt{\frac {9^j\log n}{9^{j+1}\log n}} = \frac 13,\]
and thus by \eqref{eqn:aug1}, $R_{j+1}\leq R_j$.
\hfill \halmos

Recall that in \Alg~\ref{alg:cm}, to prevent the price from increasing, we select \[P_{j+1} = \min\{P_j, \tilde P_{j+1}\}, \quad \text{where}\quad \tilde P_{j+1} = \max\{p^*(\theta): |\theta-\hat\theta_j| \leq w_j\}.\]
We next show that if the good event occurs, then this truncation is redundant, i.e., $P_{j+1}= \tilde P_{j+1}$.

\begin{proposition}[UCB is lower than the current price]\label{prop:raw}
On the event $\cal E$, we have $P_{j+1}= \tilde P_{j+1}$ for every $j\in [m]$. 
\end{proposition}
\proof{Proof.} 
We show by induction that for any $j=0, \dots, m$, we have (i) $P_j = \max\{p^*(\theta): |\theta-\hat\theta_j| \leq w_j\}$ and (ii) $\tilde P_{j+1}\leq P_j$. Note that (ii) implies that the policy selects $P_{j+1} = \tilde P_{j+1}$, which implies our claim.

For $j=0$, we have $w_j = 1$, so this is trivially true. 
Now, let us assume this holds for some $j\ge 1$. We aim to show that $\tilde P_{j+1} \leq P_j$.  
By Lemma~\ref{lem:nested}, we have
\[ \max\{p^*(\theta): \theta\in I_{j+1}\} \leq \max\{p^*(\theta): \theta\in I_{j}\}.\] 
Further, note that by definition, we have $\tilde P_{j+1}= \max\{p^*(\theta): \theta\in I_{j+1}\}$ and by \ih, we have $P_j=\max\{p^*(\theta): \theta\in I_{j}\}$. 
Combining, we obtain $\tilde P_{j+1} \leq P_j$.\hfill  \halmos 

Now we are ready to analyze the regret of the CM policy for $0$-crossing families. 
We will provide a unified analysis for the cases $s=2$ and $s>2$.

\noindent{\bf Proof of Theorem~\ref{thm:ub0d} and Theorem \ref{thm:ub0d_s}.}
\Sps\ the bad event $\mathcal{E}^c$ occurs.  
First, we show that the estimated price is close to $p^*(\theta^*)$. 
By the definition of robust \pmtn, we have 
\begin{align*}
    \|\wh \theta_{j} - \theta^*\| & = \|\Phi_{P_j}^{-1}(\bar D_j) - \Phi_{P_j}^{-1}\left(\Phi_{P_j}(\theta^*)\right)\| \\
    & \le c_2 \cdot \|\bar D_j - \Phi_{P_j}(\theta^*)\| \\
    & = c_2 \cdot \|\bar D_j  - d(P_j; \theta^*) \| \\
    & \le 2c_2 \cdot c_{\rm sg} \sqrt{\frac{\log n}{t_j}}.
\end{align*}

We next analyze $|P_{j+1}-p^*(\theta^*)|$. 
Note that by Proposition \ref{prop:raw}, we have \[P_j = \tilde P_j = \max\{p^*(\theta):\theta\in I_j\}.\]
By Assumption~\ref{assu:well-behaved}, the mapping $p^*$ is $c^*$-Lipschitz for some constant $c^*>0$, so the price $P_{j+1}$ selected in the $(j+1)$-st phase \sats\
\[|P_{j+1}-p^*(\theta^*)|\leq c^* \|\wh \theta_{j} - \theta^*\| \leq 2c_2 \cdot c^* \cdot c_{\rm sg} \sqrt{\frac{\log n}{t_j}}.\]
Since there are $t_{j+1}$ rounds in  phase $j+1$, the regret regret incurred in phase $j+1$ bounded by \[c_s \left(2c^* \cdot c_2 \cdot c_{\rm sg}  \sqrt{\frac{\log n}{t_j}}\right)^s \cdot t_{j+1}.\]
Note that there are 
$m \le \log n - \log\log n$ phases, so we can bound the cumulative regret as 
\begin{align}\label{eqn:oct1}
\mathrm{Reg}(\mathrm{CM}, \mathcal{F}) \le \sum_{j=1}^m c_s \left(2c^* \cdot c_2\cdot c_{\rm sg} \sqrt{\frac{\log n}{t_j}}\right)^s \cdot t_{j+1} = c_s \left(2c^* \cdot c_2 \cdot c_{\rm sg} \sqrt{\log n}\right)^s \cdot \sum_{j=0}^m \frac {t_{j+1}}{t_j^{s/2}}
\end{align}
We substitute $t_j$ with $\lceil 9^j \log n \rceil $ and simplify the above for $s=2$ and $s>2$ \sep ely.
When $s=2$, 
\begin{align*}
\eqref{eqn:oct1} 
&= c_s\left(2c^* \cdot c_2 \cdot c_{\rm sg} \sqrt{\log n}\right)^2 \cdot \sum_{j=0}^K \frac {t_{j+1}}{t_j} 
\\
&\leq c_s\left(2c^* \cdot c_2\cdot c_{\rm sg}\right)^2 \cdot \log n\cdot 9(\log n -\log\log n)\\
&=O(\log^2 n),
\end{align*}
and we obtain Theorem \ref{thm:ub0d}. 
Now \sps\ $s>2$. Then, 
\begin{align*}
\eqref{eqn:oct1} &= c_s\left(2c^*\cdot c_2 \cdot c_{\rm sg} \sqrt{\log n}\right)^s \sum_{j=0}^m \frac {9^{j+1}\log n}{9^{j\cdot \frac s2} \log^{s/2} n}\\
&\le c_s \left(2c^* \cdot c_2 \cdot c_{\rm sg}\right)^s \log n \cdot \sum_{j=0}^m 9^{(1-\frac s2)j +1}\\
&\leq 2c_s \cdot \left(2c_2\cdot c^*\cdot c_{\rm sg} \right)^s \log n \cdot \int_0^m 9^{(1-\frac s2)x} dx\\
&= 2c_s \cdot \left(2c^* \cdot c_2 \cdot c_{\rm sg}\right)^s \log n \cdot \frac {2}{(s-2) \log 9}\\
&= O(\log n), 
\end{align*}
and we obtain Theorem~\ref{thm:ub0d_s}. \hfill\halmos}

\subsection{Finite-crossing Family}
In this section, we show that the ICM policy has $\tilde O(n^{k/(k+1)})$ regret, as stated in Theorem \ref{thm:ubgen}.

\subsubsection{Clean events.}
As in the $0$-crossing case, to simplify the exposition, we start by defining the clean events. 
Recall that $[L_j,U_j]$ is the \ci\ computed at the end of phase $j$. 

\bdefn[Clean events]  
For each $j\in [m]$, let $\mathcal{E}_j$ be the event that $p^*(\theta^*) \in [L_j,U_j]$.
\edefn

\begin{lemma}[Clean events are likely]
\label{lem:clean_event}
For any $j\in [m]$, we have $\ho{P}(\mathcal{E}_j)\geq 1- (k+1) n^{-2}$.
\end{lemma}
\proof{Proof.} 
For each $i=0,\dots,k$, denote by $\bar D_i$ the \emp\ mean demand at the exploration price $P_j - i h$. 
Denote $\Phi:=\Phi_{P_i,P_i-h,\dots,P_i-kh}$ and ${\bf \bar D} := (\bar D_0,\dots,\bar D_k)$. Note that under this notation, the estimated \pmt\ is $\Phi^{-1}({\bf D})$, so our goal is essentially to bound the difference between $\Phi^{-1}({\bf D})$ and $\theta^* = \Phi^{-1}(\Phi(\theta^*))$.

By Hoeffding's \ineq\ (Theorem~\ref{thm:hoeffding}), for each $i=0,1,\dots,k$, w.p. $1- (k+1)n^{-2}$ we have  
\begin{align*}
\lb|d(P_j - ih; \theta^*) - \bar D_i \rb|\le 2c_{\rm sg}\cdot \sqrt \frac {\log n}{n_j}.
\end{align*}
By the union bound, it then follows that w.p. $1-kn^{-2}$ we have 
\[\| \Phi(\theta^*) - {\bf \bar D}\|_\infty \le 2c_{\rm sg}\cdot \sqrt \frac {\log n}{n_j}.\]
\Sps\ the above \ineq\ holds. Then, by the definition of crossing number, we have  
\begin{align*}
\|\theta^* - \wh \theta\|_1
&= \|\Phi^{-1}({\bf \bar D}) - \Phi^{-1}(\Phi(\theta^*))\|_1 \\
&\leq c_2 h^{-k} \cdot \|\Phi(\theta^*) - {\bf \bar D}\|_1\\
&\le c_2 h^{-k} \cdot \|\Phi(\theta^*)-{\bf \bar D}\|_\infty \cdot \sqrt{k+1}\\
&\le c_2 h^{-k} \cdot c_{\rm sg} \cdot 2\sqrt\frac {\log n}{n_j}\cdot \sqrt{k+1}.
\end{align*}
By Lemma~\ref{lem:clean_event} and the Lipschitzness (see Assumption \ref{assu:well-behaved}) of the function  $p^*(\cdot)$, we deduce that 
\[|p^*(\wh\theta) - p^*(\theta^*)| \le c^* \|\theta^* - \wh \theta\|_1 \le c^* c_2 h^{-k} \cdot c_{\rm sg} \cdot 2\sqrt\frac {\log n}{n_j}\cdot \sqrt{k+1} = w_j.\]
Recall that $L_j = p^*(\wh \theta) - w_j$ and $U_j = p^*(\wh \theta) + w_j$, so we conclude that $p^*(\theta^*)\in [L_j,U_j]$.\hfill\halmos

\subsubsection{Regret analysis.}
Recall that in each phase $j\in [m]$, we select each of these $k+1$ exploration prices $n_j$ times.
Furthermore, there are at most $m$ exploration phases and may stop exploration early if the  the current price is lower than the left boundary of the \ci. 
we will consider the phase after which the policy stops exploration, formally defined below.

\bdefn[Termination of exploration]\label{def:last_exploration}
For each $\ell\in [m]$, define $n^\ell:=\sum_{j=1}^\ell n_j\cdot (k+1)$.
The {\em last exploration phase} is a \rv\ given by \[J:= \min\lb\{j\in \{1,\dots,m\}: P_j-kh <L_j \ \text{or} \  j=m\rb\}.\]
We also define \[S:= \sum_{j=1}^J (k+1)\cdot n_j.\]
\edefn

The ``or'' part in the above ensures that the set is not empty.
We next bound the regret incurred in the exploration phases, i.e., until time $S$.

\begin{lemma}[Exploration regret]\label{lem:exploration_regret} Denote by $(X_t)$ the random price sequence selected by the ICM policy. 
\Sps\ that the events $\mathcal{E}_1,\dots, \mathcal{E}_m$ occur. 
Then, for any $t$ with $n^{j-1}\le t\le n^j$ for some  $j\le J$, it holds that \[r^* - r(X_t) \le c_s  \left(c^* w_{j-1}\right)^s.\]
Furthermore, for any $t\ge n^J = S$, it holds that 
\[r^* - r(X_t) \le c_s \cdot (c^* w_m)^s.\]
\end{lemma}
\proof{Proof.}
Fix any phase $j\le J, i\in \{0,\dots,k\}$, and consider the price $P_j - i\cdot h$.
Since $P_j \le U_{j-1}$, we have \begin{align}\label{eqn:121623a} P_j-ih\in [L_{j-1}, U_{j-1}].
\end{align}
By the definition of $(\mathcal{E}_j)$, we have
\begin{align}\label{eqn:121623b} L_{j-1} \le p^*(\theta^*) \le U_{j-1},\end{align}
and hence 
\[\lb|(P_j- ih) - p^*(\theta^*)\rb| \le U_{j-1}-L_{j-1} \le c^* w_{j-1}.\]
Finally, by the definition of $s$-sensitivity and note that for every round $t$ in phase $j$, it holds that $X_t \in \{P_j, P_j - h, \dots, P_j-kh\}$ a.s., we conclude that
\[r^* - r(X_S)\le c_s \cdot \left(c^* w_{j-1} \right)^s.\eqno\halmos\]

Next, we turn to the regret in the exploitation phase.

\begin{lemma}[Exploitation regret] \Sps\ the clean event holds, then \[r^* - r(X_S) \le c_s (m(k+1)h)^s.\]
\end{lemma}
\proof{Proof.} Consider the last phase $I$ where the good event occurs, formally, 
\[I := \max\{i=\{0,1,\dots,m\}: P_i-kh \ge U_i\},\]
which is well-defined since the $i=0$ \sats\ the \ineq\ and hence the set is non-empty.
By the definition of $J$, we have 
\begin{align}\label{eqn:121723a}
P_J - kh < L_J.    
\end{align}
By Lemma \ref{lem:clean_event}, if the clean events occur, then
\begin{align}\label{eqn:121723b}
L_J\le p^*(\theta^*)\le U_J,
\end{align}
and moreover, for any $i<j$, we have $[L_j, U_j]\subset [L_i, U_i]$, and hence
\begin{align}\label{eqn:121723c}
U_J \le U_I.
\end{align} 
Finally, dote that $I\le J-1$ a.s., so by the construction of the ICM policy, we have 
\begin{align}\label{eqn:121723d}
|U_I - P_J| \le (J-I+1)\cdot (k+1)h.
\end{align} 
Combining eqn. \eqref{eqn:121723a},\eqref{eqn:121723b},\eqref{eqn:121723c} and \eqref{eqn:121723d}, we obtain  
\begin{align*}
|P_J - kh - p^*(\theta^*)| &\leq |P_J - kh - U_I|\\
&= |P_J -kh - P_J| +|P_J - U_I| \\
&\le (J-I+1)\cdot (k+1)h\\
&\le m(k+1)h.
\end{align*}
Therefore, since the revnue function is $s$-sensitive, we conclude that
\[r^* - r(X_S) \le c_s \cdot (m(k+1)h)^s.\eqno\halmos\] 

\subsubsection{Proof of Proposition~\ref{prop:ub_gen_d}.}
First we decompose the regret as follows:
\begin{align}\label{eqn:121723e}
\mathrm{Reg}(\mathrm{ICM},\mathcal{F}) 
&= \ho{E}\lb[\sum_{t=1}^n \lb(r^*-r(X_t)\rb)\Big | \bigcap_{j=1}^J \mathcal{E}_j \rb] + n\cdot \ho{P}\lb[\bigcup_{j\in [m]}\overline{\mathcal{E}_j}\rb]\notag \\
&\le \sum_{j=1}^J \ho{E}\lb[\sum_{t=n^{j-1}}^{n^j} \lb(r^* - r(X_t)\rb)\Big |\bigcap_{j\in [m]}\mathcal{E}_j\rb] +\ho{E}\lb[\sum_{t=n^J}^n \lb(r^* - r\lb(X_S\rb)\rb)\rb] + n\cdot n^{-2}\cdot m,
\end{align}
where the \ineq\ follows from Lemma \ref{lem:clean_event}.
Recall from Definition \ref{def:last_exploration} that $S = n^J$, so by Lemma \ref{lem:exploration_regret}, we have  
\begin{align*}
\eqref{eqn:121723e} &\le n_1 + \sum_{j=2}^m c_s \left(c^* w_j\right)^s n_j +  c_s\left(c^* w_m\right)^s n + c_s (m(k+1)h)^s \cdot n + \frac{m(k+1)}{n^2}\\
&\le n_1 + \sum_{j=2}^m c_s  \left(2c^* c_{\rm sg} h^{-k} \sqrt \frac {2c_5 k\log n}{n_{j-1}}\right)^s n_j + c_s \left(2c^* c_{\rm sg} h^{-k} \sqrt \frac {2c_5 k\log n}{n_m}\right)^s n + c_s (2mkh)^s n + \frac{2mk}{n^2}\\
& = n_1 + c_s \left(4c^* c_{\rm sg} h^{-k}\sqrt{c_5 k \log n}\right)^s  \left(\sum_{j=1}^{m-1}  n_{j-1}^{-s/2} n_j + n_m^{-s/2} n\right) + c_s (2mkh)^s n + \frac{2mk}{n^2}.
\tag*\halmos
\end{align*}

\subsubsection{Proof of Theorem \ref{thm:ub_gen_s}: Choosing the policy parameter.}
Now, let us ignore the $O(\log n)$ terms in the regret bound in Proposition \ref{prop:ub_gen_d} and determine the \pmt s to minimize the \asym\ order, thus proving Theorem \ref{thm:ub_gen_s}.
To simplify, we view $k,m$ as small constants relative to $n$ and ignore them.

To further simplify the notation, we write $h=n^{-y}$ and $n_i=n^{z_i}$. 
Under this notation, for any $j=0,1,\dots,m-1$, the expected regret in the $(k+1)$-st exploration phase is on the order of 
\[\lb(h^{-k}n_j^{-1/2}\rb)^s n_{j+1} = n^{sk y-\frac s2 z_j+z_{j+1}}.\]
On the other hand, from in eqn. \eqref{eqn:121223}, we notice that the regret in the exploitation phase is \asymly\ 
\[h^{-k}n_m^{-2/s} n + h^s n,\]
ignoring $k,m,\log n$  terms and constants. 
These two terms can be interpreted as the regret caused by (i) estimation error and (ii) overshooting the optimal price \resp.
Under our notation, we can simplify them as 
\[h^{-k} n_m^{-s/2} n =  n^{sky + 1 - \frac s2 z_m}\quad \text{and}\quad h^s n = n^{1-sy}.\]
Therefore, we have reduced the problem of finding the optimal \pmt s  to the following LP:
\begin{equation}
\begin{aligned}
\mathrm{LP}(d):\quad  &\min_{x,y,z}\quad  \quad & n^x \\
&\textrm{subject to}  & n^{z_1} &\leq n^x, &\text{Regret in phase }1\\
& & n^{2sk y+z_2 - \frac s2 z_1} & \leq n^x, &\text{Regret in phase }2\\
& &\dots\\
& & n^{sky + 1 - \frac s2 z_{m-1}} & \leq n^x, &\quad \quad \text{Regret in phase } m\\
& & n^{sky + 1 - \frac s2 z_m} & \leq n^x, &\quad \quad \text{Regret in exploitation due to estimation error}\\
& & n^{1-sy}&\leq n^x, &\quad \text{Regret in exploitation due to overshooting}\\
& & x,y,z\geq 0, &\quad z\leq 1\notag
\end{aligned}
\end{equation}

Taking logarithm with base $n$ on both sides, the above becomes 
\begin{align*}
\min_{x,y,z} &\quad \quad x\\  
\textrm{s.t.} & \quad
\begin{bmatrix}
-1& 0 &1 & 0& 0 &0 & \cdots &0\\
-1& sk &-\frac s2 &1 &0 &0 &\cdots &0\\
-1& sk &0 &-\frac s2 &1 &0 & \cdots &0\\
-1& sk &0 &0 &-\frac s2 & 1 &\cdots &0\\
& &&\cdots \cdots&&\\
-1& sk &0 &0 &0 &\cdots&-\frac s2 & 1\\
-1& sk &0 &0 &0 &0 &\cdots &-\frac s2\\
-1& -s &0 &0 &0 &0 &\cdots&0\\
\end{bmatrix}
\begin{bmatrix}
x\\
y\\
z_1\\
z_2\\
\vdots\\
z_{m-1}\\
z_m
\end{bmatrix}
\leq
\begin{bmatrix}
0\\
0\\
\vdots\\
0\\
-1\\
-1
\end{bmatrix}\\
&x,y,z \ge 0,  \\
&z \le 1.
\end{align*}


Note that this LP consists of $m+2$ variables and $m+2$ \ineq\ \constr s, so the minimum is attained when all \ineqs\ become identities, i.e.,
\begin{align}
z_1 &= x \label{eqn:may18}\\
z_2 - \frac s2 z_1 &= x-sky\notag\\
z_3 - \frac s2 z_2 &= x-sky\notag\\
\dots\notag\\
z_m - \frac s2 z_{m-1} &= x-sky\notag\\
1-\frac s2 z_m &= x-sky \notag\\
1-sy &= x\label{eqn:oct4}
\end{align}
By telescoping sum, we have 
\[1-\left(\frac s2\right)^m z_1 = \left(1+\frac s2 +\dots + \left(\frac s2\right)^{m-1}\right) \left(x-sky\right).\]
Combining the above with eqn. \eqref{eqn:may18} and \eqref{eqn:oct4}, we obtain
\[1+\left(1+\frac s2 + \dots + \left(\frac s2\right)^{m-1}\right) k(1-x) = \left(1+\frac s2 +\dots + \left(\frac s2\right)^m\right) x.\]
Rearranging, we have
\[x=\frac{1+\left(1+\frac s2 + \dots +\lb(\frac s2\rb)^{m-1}\right)k}{\left(1+\frac s2 + \dots +\lb(\frac s2\rb)^{m-1}\right) (k+1) + \left(\frac s2\right)^m.}\]
In \parti, for $s=2$, the above becomes 
\[x= \frac{mk+1}{m(k+1)+1} = \frac k{k+1} + \frac 1{m(k+1)^2.}\]
Note for any $n$ it holds that $n^{1/\log n}=O(1)$, thus for $m=\log n$, we have \[{\rm Reg}\lb({\rm ICM}, \mathcal{F}\rb) = \tilde O\lb(m \cdot n^x\rb) = \tilde O\lb(n^{\frac k{k+1}}\rb). \eqno\halmos\]

\iffalse
\subsection{Infinite-crossing Family}
In this section we first present a general regret upper bound for policy $\mathrm{UE}_{\Delta,w}$, which immediately implies
Theorem~\ref{thm:ub_infty}. 
To this aim, we need to introduce another constant $\eta$, as motivated by the  the following result.
For notational convenience, we abbreviate $\frac {\partial^k} {\partial x^k} r(x;\theta)$ as $R^{(k)}(x;\theta)$ for any $k\geq 0$.
\begin{lemma}\label{lem:dec18}
Let $\mathcal{F}=\{r(x,\theta):\theta\in \Theta\}$ be a family of $s$-sensitive reward \func s. 
Then, there exists a constant $\eta>0$ such that for any $\theta\in \Theta$ and $x\in \left[p^*(\theta) - \eta, p^*(\theta)\right],$ it holds $R^{(s)}(x;\theta)<0$ and
\[2 R^{(s)}(p^*(\theta);\theta)\leq  R^{(s)}(x;\theta) \leq \frac 12 R^{(s)}(p^*(\theta);\theta).\]
\end{lemma}

\proof{Proof.}
First consider any fixed $\theta\in \Theta$.
By definition of sensitivity, we have $R^{(1)}(p^*(\theta),\theta)=...=R^{(s-1)}(p^*(\theta),\theta)=0$ and $R^{(s)}(p^*(\theta),\theta)<0$.
Define
\[g(\theta)=\sup \left\{\gamma\geq 0\ |\ 2 R^{(s)}(p^*(\theta);\theta)\leq  R^{(s)}(x;\theta) \leq \frac 12 R^{(s)}(p^*(\theta);\theta),\quad \forall x\in \left[p^*(\theta) - \gamma, p^*(\theta)\right]\right\}.\]
By continuity of $R^{(s)}$ in $x$, we have $g(\theta)>0$ for any $\theta\in \Theta$. 
We complete the proof by showing that $\eta := \suP_{\theta\in \Theta} g(\theta)>0$.
Recall that $\Theta$ is compact, and $R^{(s)}$ is continuous in $\theta$, we know that $\eta$ can be attained, i.e., there exists some $\theta\in \Theta$ with $g(\theta)=\eta$.
Moreover, note that for any $\theta$ we have $g(\theta)>0$, therefore $\eta>0$, and the proof completes.
\halmos

We are now ready to state the main result in this section.
Note that by choosing $\Delta=n^{-1/(3s+1)}$ and $w=n^{-2/(3s+1)}$, we immediately obtain Theorem~\ref{thm:ub_infty}.
\begin{proposition}[Upper Bound]\label{prop:5}
Let $\mathcal{F}$ be any $s$-sensitive family for some $s\geq 2$.
\Sps\ $\Delta \leq \frac {c_s}{8s!  C^{(1)}}\eta^s$ and  $m\geq 4$, then
$$\mathrm{Reg}\left(\mathrm{UE}_{\Delta,m}, \mathcal{F} \right)= O\left(\Delta^{-1} w^{-2}\log n + (w+\Delta^s) T\right)$$
where we recall that $w=2c_{\rm sg} \sqrt{\frac{\log n}m}$.
\end{proposition}
Our analysis proceeds by conditioning on the following the notion of clean event, which occurs with high \prb\ as we will show soon.
\bdefn[Clean event]
Let $\mathcal{E}_j$ be the event that $\big |r(x_j) - \bar \mu_j\big| \leq 2c_{\rm sg} \sqrt{\frac{\log n}m}$, and $\mathcal{E} = \bigcaP_{j=1}^{\lceil \Delta^{-1}\rceil} \mathcal{E}_j$.
\edefn
Note that by our choice of $L_j,U_j$, we know that $\mathcal{E}$ is simply the event that $r(x_j)\in [L_j,U_j]$ for all $1\leq j\leq \Delta^{-1}$.
We next show that $\mathcal{E}$ occurs with high \prb, and hence we may perform the analysis conditional on $\mathcal{E}$.
\begin{lemma}
$\ho{P}(\overline{\mathcal{E}})\leq n^{-1}$.
\end{lemma}
\proof{Proof.}
Let $R$ be the true reward \func.
Recall that $Z_i^j$ for $i=t^{(j-1)}+1,..., t^{(j)}$ are i.i.d. samples from a subgaussian \distr\ with mean $r(x_j)$, and that Assumption~\ref{assu:subg} the sugaussian norm of this \distr\ is at most $c_{\rm sg}$.
Thus by Lemma~\ref{lem:ci_0dim}, we have $\ho{P}[\overline{\mathcal{E}_j}]\leq n^{-2}$.
By the union bound, we have
$\ho{P}[\mathcal{E}] \geq 1-n^{-2} \cdot \log n \geq 1-n^{-1}.$

In the rest of this section we will fix a true reward \func\ $r(x;\theta)$ and write $x^* = p^*(\theta)$ and $r(x)=r(x;\theta)$.
\bdefn
Define $x_\ell$ be the closest sample price to $x^*$, i.e. 
$\ell := \arg\min_{0\leq j\leq \Delta^{-1}} \{|x_j - x^*|\}.$
\edefn
We first show that conditional on $\mathcal{E}$, the policy will stop reducing the price and enter the exploitation phase before reaching $x^*-\eta$.
\begin{lemma}\label{lem:dec28}
\Sps\ $\mathcal{E}$ occurs. For any $m\geq  \lb(\frac{4s!\cdot8\cdot c_{\rm sg}}{c_s}\rb)^2 \cdot \eta^{-2s} \log n$ and $\Delta \leq \frac {c_s}{8s! C^{(1)}}\eta^s$, we have $x_h \geq x^*-\eta$.
\end{lemma}
\proof{Proof.}
Recall that $x$ is said to be a sample price if $x=1-j\Delta$ for some integer $j$.
Consider the smallest sample price $\tilde x$ above $x^*-\eta$, then $|x^*-\eta - \tilde x|\leq \Delta$. 
By Assumption~\ref{assu:smooth}, the first derivatives are bounded by $C^{(1)}$ and hence $R$ is 
$C^{(1)}$-Lipschitz, so
\[|r(\tilde x) - r(x^*-\eta)|\leq  C^{(1)}|x^*-\eta - \tilde x| \leq  C^{(1)}\Delta \quad \text {and} \quad |r(x_\ell) - r(x^*)|\leq C^{(1)}\Delta.\]
Moreover, by Theorem~\ref{thm:lag}, and since $R^{(1)}(x^*)=...=R^{(s-1)}(x^*)=0$, 
\begin{align}\label{eqn:may21c}
    |r(x^*-\eta) -r(x^*)| = \left|\frac {R^{(s)}(\xi)}{s!}\eta^s\right|
\end{align}    
for some $\xi \in (x^*-\eta, x^*)$.
By Lemma~\ref{lem:dec18}, $|R^{(s)}(\xi)| \geq \frac 12 \cdot |R^{(s)}(x^*)|$, so 
\begin{align}\label{eqn:may21d}
    |r(x^*-\eta)- r(x^*)| \geq \frac{|R^{(s)}(x^*)|}{2s!}\eta^s.
\end{align}
By combining the \ineqs\ \eqref{eqn:may21c} and \eqref{eqn:may21d}, we have 
\[r(\tilde x) \leq r(x_\ell) - \left(\frac {|R^{(s)}(x^*)|}{2s!}\eta^s - 2C^{(1)}\Delta\right).\]
Recall that $|R^{(s)}(x^*)|\geq c_s$, so for any $\Delta \leq \frac {c_s}{8s! C^{(1)}}\eta^s$, we have 
\begin{align}\label{eqn:may21}
\frac {|R^{(s)}(x^*)|}{2s!}\eta^s - 2C^{(1)}\Delta \geq \frac {|R^{(s)}(x^*)|}{4s!}\eta^s.
\end{align}
Hence, \sps\ $m\geq  \lb(\frac{4s!\cdot8\cdot c_{\rm sg}}{c_s}\rb)^2 \cdot \eta^{-2s} \log n$, then $4w\leq \frac {c_s}{4s!}\eta^s\leq \frac {|R^{(s)}(x^*)|}{4s!}\eta^s$, and by \eqref{eqn:may21}
\begin{align}\label{eqn:may21b}
r(\tilde x)< r(x_\ell) - 4w.
\end{align}
Since $\mathcal{E}$ occurs, we have 
$|U(\tilde x) - r(\tilde x)| \leq w$ and $|L(x_\ell) - r(x_\ell)| \leq w$. Combining with \eqref{eqn:may21b}, we obtain $U(\tilde x) < L(x_\ell)$,
and thus the halting criterion is \satd\ at $\tilde x$, so $x_h\geq x^*-\eta$.

\begin{lemma}\label{lem:may23a}
\Sps\ $x_{k+\ell}\geq x^* - \eta$. Then, \[|r(x_{k+\ell}) - r(x^*)|\geq \frac{{R^{(s)}}(x^*)}{2s!}\lb(\lb(k-1\rb)\Delta\rb)^s.\]
\end{lemma}
\proof{Proof.}
By Theorem~\ref{thm:lag}, $|r(x_{k+\ell}) - r(x^*)| = \lb|\frac1{s!} R^{(s)}(\xi)\cdot (x_{k+\ell} - x^*)^s\rb| \geq \frac1{2s!} |R^{(s)}(x^*)| \cdot |x_{k+\ell} - x^*|^s.$
By definition of $x_\ell$, we have $|x_\ell - x^*|\leq \Delta$, so 
$|x^*-x_{k+\ell}| \geq |x_\ell - x_{k+\ell}| - |x^* - x_\ell| \geq (k-1)\Delta.$ 
Thus, $|r(x_{k+\ell}) - r(x^*)|\geq \frac1{2s!} |R^{(s)}(x^*)| \cdot \lb(\lb(k-1\rb)\Delta\rb)^s.$

\begin{lemma}\label{lem:may23b}
\Sps\ $\mathcal{E}$ occurs and $i:= \arg\max_{0\leq j\leq \Delta^{-1}} \{L_j\}$. 
Then, 
\[|r(x_i) - r(x_\ell)|\leq \max\{2w,\frac{2R^{(s)}(x^*)}{s!} \Delta^s\}.\]
\end{lemma}
\proof{Proof.}
Since $|x_{\ell}-x^*|\leq \Delta \leq \eta$, by Theorem~\ref{thm:lag}, it holds that $r(x_\ell)\geq r(x^*) - \frac{2R^{(s)}}{s!} |x^* - x_\ell|^s\geq r(x_i) - \frac{2R^{(s)}}{s!} |x^* - x_\ell|^s$, and thus
$r(x_i)-r(x_\ell) \leq \frac{2R^{(s)}}{s!} |x^* - x_\ell|^s \leq \frac{2R^{(s)}}{s!} \Delta^s$.
\OTOH, by definition of $i$, we have $L_\ell \leq L_i$. 
Since $\mathcal{E}$ occurs, we also have $|L_i - r(x_i)| \leq w$ and $|L_\ell - r(x_\ell)| \leq w$, and therefore
$r(x_\ell) - r(x_i)\leq 2w$, and the proof follows.
\halmos

\begin{figure}[h]
\centering
\includegraphics[width=8cm]{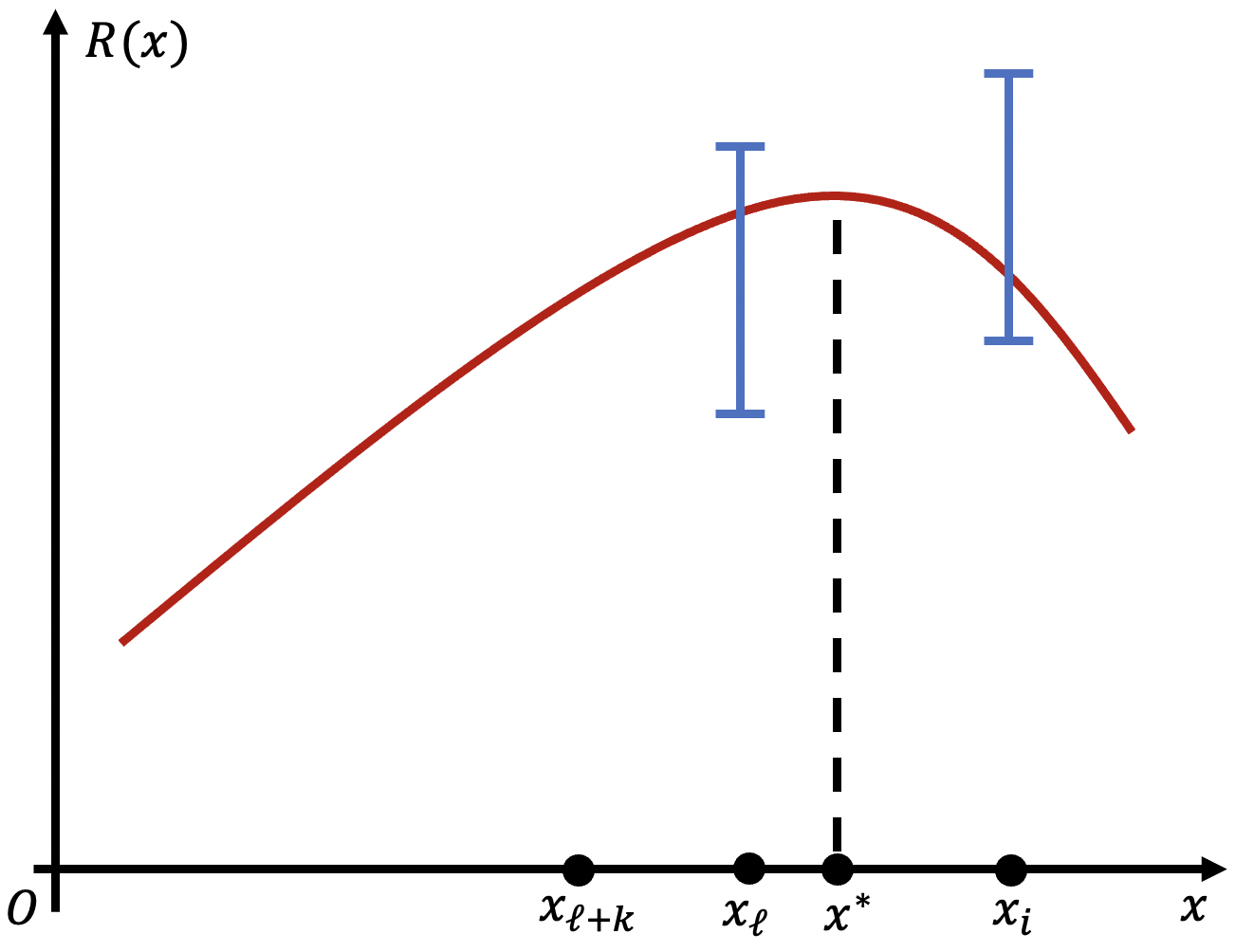}
\caption{Illustration of Lemma~\ref{lem:dec21}}
\label{fig:6delta}
\end{figure}

\begin{lemma}\label{lem:newton_leibniz}
\Sps\ $\mathcal{E}$ occurs and $\Delta\leq \eta$. 
Then for any $k\geq 3$ with $x_{k+\ell}\geq x^*-\eta$, \[|r(x_{\ell+k}) - r(x_i)| \geq \frac{c_s}{2^s s!}(k\Delta)^s - 2w.\]
\end{lemma}
\proof{Proof.}
By Lemma~\ref{lem:may23a}, \ref{lem:may23b} and the triangle \ineq,
\begin{align*}
|r(x_{\ell+k}) - r(x_i)| &\geq |r(x_{\ell+k}) - r(x^*)| - |r(x^*) - r(x_\ell)| - |r(x_i) - r(x_\ell)| \\ 
&\geq \frac C{s!} \cdot \left(\left(k-1\right)\Delta\right)^s - \frac{c_s}{s!}\Delta^s - 4w + \frac{c_s}{s!}\Delta^s\\
&\geq \frac{c_s}{s!}\Delta^s \left((k-1)^s - 1\right)- 4w\\
&\geq \frac{c_s}{2^s s!}(k\Delta)^s - 4w,
\end{align*}
and the proof follows.
\halmos

The crux for proving our lower bound lies in analyzing the regret in the exploitation phase. 
To this aim, we use the above lemma to bound the per-round regret in the exploitation phase, formally stated below, and illustrated in Figure \ref{fig:6delta}. 
Recall that $x_h$ is the halting price, which is selected in every round in the exploitation phase.
\begin{lemma}\label{lem:dec21}
\Sps\ $\mathcal{E}$ occurs, then
$r(x_h) - r(x^*) \leq \frac{3^{s} s! \cdot 4w}{c_s} + \max\{3^s, c_s\}\cdot \Delta^s.$
\end{lemma}
\proof{Proof.}
Consider two cases regarding $h$ and $\ell$.

\noindent{\bf Case 1.} 
\Sps\ $h\geq \ell-2$, i.e.
$x_h\geq x_\ell- 2\Delta$. 
Since $|x_\ell - x^*|\leq \Delta$, we have 
$|x_h - x^*| \leq |x_\ell - x_h| + |x^* - x_\ell| \leq 2\Delta + \Delta = 3\Delta.$
Thus by definition of sensitivity, when $|x_h - x^*|\leq \eta$ it holds
$|r(x^*) - r(x_h)| \leq c_s \cdot  |x_h - x^*|^s \leq c_s \cdot 3^s \Delta^s.$

\noindent{\bf Case 2.} 
Now \sps\ $h\leq \ell-3$, i.e. $x_h\leq x_\ell - 3\Delta$. 
Let $k = h-\ell-1$, so that  $x_{\ell+k}$ is the last sample price that the UE policy selected before halting at $x_h$.
Then by definition of $x_h$, the halting criterion is {\it not} \satd\ at the $x_{\ell+k}$, i.e. $[L(x_i),U(x_i)] \cap [L(x_{\ell+k}), U(x_{\ell+k})]\neq \emptyset$, so 
$|r(x_i)- r(x_{\ell + k}) | \leq 4w.$
Combining with  Lemma~\ref{lem:newton_leibniz}, we have $2w \geq |r(x_i) - r(x_{\ell + k})| \geq \frac{c_s}{2^s s!}(k\Delta)^s - 2w,$
i.e.,
\begin{align}\label{eqn:apr10}
(k\Delta)^s \leq \frac{2^s s!\cdot 4w}{c_s}.
\end{align}
Note that $|x_h - x^*|\leq (k+1)\Delta$, and recall that conditional on $\mathcal{E}$, we have $x_h\geq x^*-\eta$, \ift\
\begin{align*}
|r(x_h) - r(x^*)|&\leq
c_s\cdot \left((k+1)\Delta\right)^s &\text{by Theorem~\ref{thm:lag}}\\
&\leq c_s\cdot \left(\frac 32 k\Delta\right)^s &\text{since } k=h-\ell-1\geq 2 \\
&\leq \frac{4w \cdot 3^{s} s!}{c_s} 
= \frac{4\cdot 3^{s} s! \cdot w}{c_s}, &\text{by \eqref{eqn:apr10}}
\end{align*}
and the proof is complete.
\halmos

\noindent{\bf Proof of Proposition~\ref{prop:5}.}
Fix any $R\in \mathcal{F}$. 
\Sps\ $\mathcal{E}$ does not occur, then the regret is at most $n$.
\Sps\ $\mathcal{E}$ occurs, 
then by Lemma~\ref{lem:dec21}, the regret incurred in the exploitation phase is bounded by $\left(\frac{6\cdot c_s \cdot 3^{s} s!}{c_6}\cdot w + \max\{3^s, c_s\}\cdot \Delta^s\right)T$.

On the other side, recall that each sample price is selected for at most $m$ times, so the cumulative regret incurred in the exploration phase is bounded by $mT$.
Moreover, there are at most $\lceil\Delta^{-1}\rceil$ sample prices.
Recalling that $w = 2c_{\rm sg} \sqrt{\frac {\log n}m}$, i.e. 
$m=4c_{\rm sg}^2 w^{-2}\log n$, we can bound the total regret as  
\begin{align*}
&\quad \mathrm{Reg}(\mathrm{UE}_{\Delta,w},R) \\
&\leq \ho{P}[\mathcal {\overline E}] \cdot T + \ho{P}[\mathcal{E}] \cdot \left(4c_{\rm sg}^2 w^{-2} \Delta^{-1}\log n + \left(\frac{6\cdot c_s \cdot 3^{s} s!}{c_6}\cdot w + \max\{3^s, c_s\}\cdot \Delta^s\right)\cdot T \right)\\
&\leq n^{-1}\cdot T + \left(4c_{\rm sg}^2 w^{-2} \Delta^{-1}\log n + \left(\frac{6\cdot c_s \cdot 3^{s} s!}{c_6}\cdot w + \max\{3^s, c_s\}\cdot \Delta^s\right)\right)\cdot T\\
&= O\left(\Delta^{-1} w^{-2}\log n + (\Delta^s + w)T\right),
\end{align*}
and Proposition~\ref{prop:5} follows. 
\halmos
\fi

%% file: lb.tex
{\clean We prove our lower bounds in a unified way using the following tool, due to \cite{wald1948optimum}.

\subsection{Adaptive \clf s and the Wald-Wolfowitz Theorem}
Our proof considers Bernoulli reward \distr\ at each price.
To formally define the notion of an \adap\ \clf, we need a standard concept from \prb\ theory called the {\em stopped $\sigma$-algebra}.
In words, an event is in this $\sigma$-algebra if it depends on the information until a stopping time. 

\bdefn[Stopped $\sigma$-algebra]
Let $m\ge 1$ be an integer and consider a filtered \prb\ space $(\Omg, \mathcal{A}, (\mathcal{F}_t)_{t=1}^m, \ho{P})$. 
An integer-valued  \rv\ $\tau$ is a {\em stopping time}, if for any $t\in \real$, the event $\{\tau > t\}$ is $\mathcal{F}_t$-measurable. 
The {\em stopped $\sigma$-algebra} is \[\mathcal{F}_\tau:= \{A\in \mathcal{A} \mid \{\tau>t\} \cap A \in \mathcal{F}_t,\ \forall t\ge 1\}.\]
A \rv\ $X$ is {\em $\mathcal{F}_\tau$-measurable} if for any Borel set $B\subset \real$, we have $X^{-1}(B) \in \mathcal{F}_\tau$.
\edefn

\bdefn[Adaptive Classifier]\label{def:confi_leaf_coloring}
Consider demand \func s $f_0, f_1: [0,1]\rar [0,1]$.
Let $(X_t)$ be any policy.
For $i=0,1$, denote by $\ho{P}_i$ the \prb\ measure on $([0,1]\times \real)^T$ where $D_t \sim {\rm Ber}(f_i(X_t))$ for each $t\in [T]$.
An {\it \adap\ \clf} (or simply {\em \clf}) for $f_0,f_1$ is a tuple $(I, \tau)$ where $\tau$ is a stopping time w.r.t. the filtration $(\mathcal{F}_t)$ and $I$ is an $\mathcal{F}_\tau$-measurable \rv\ taking value from $\{0,1\}$.
\edefn

We measure the quality of an \adap\ \clf\ by the \prb\ that it returns a wrong output, as formalized below.

\bdefn[Confidence of an \adap\ \clf] 
For any $\alpha,\beta\in [0,1]$, an \adap\ \clf\ $(I,\tau)$ is {\it $(\alpha, \beta)$-confident} if
\[\ho{P}_0\left(I = 0\right) \geq \alpha \quad \text{and}\quad \ho{P}_1\left(I=0\right) \le \beta.\]
\edefn

The probabilities $\ho{P}_0\left(I=1\right)$ and $\ho{P}_1\left(I=0\right)$ are usually referred to as the {\em type I} and {\em type II error} in statistics. 
It is easy to verify that an \adap\ \clf\ is $(\alpha,\beta)$-confident if and only if the type I and type II errors are (at most) $1-\alpha$ and $\beta$ \resp.

All of our lower bounds rely upon the following lower bound of the sample complexity for distinguishing between two demand models using an \adap\ \clf, due to \cite{wald1948optimum}.
It states that any \adap\ \clf\ achieving a given confidence level must query at least a certain number of samples in {\em expectation}.
Denote by $F_\mu$ the cdf of a Bernoulli \rv\ with mean $\mu\in [0,1]$.

\begin{theorem}[Sample complexity lower bound for \clf s]
\label{thm:ww}
Consider functions $f_0,f_1: [0,1]\rightarrow [0,1]$. 
For $i=0,1$, denote $\Delta(f_i,f_{1-i}) = \max_{p\in [0,1]} \mathrm{KL} \lb(F_{f_i(p)}, F_{f_{1-i}(p)}\rb).$
Then for any $(\alpha, \beta)$-\conf t \clf\ $(I,\tau)$, we have
\begin{align}
\ho{E}_0[\tau] \geq \frac{ \alpha \log \frac{\alpha}{\beta} + (1-\alpha) \log \frac{1-\alpha}{1-\beta}}{\Delta(f_0,f_1)} \quad \text{and}\quad
\ho{E}_1 [\tau] \geq \frac{ \beta \log \frac{\beta}{\alpha} + (1-\beta) \log \frac{1-\beta}{1-\alpha}}{\Delta(f_1,f_0)}.
\end{align}
\end{theorem}

For example, when $\alpha=\frac 34, \beta=\frac14$, the above becomes $1/\Delta(f_0,f_1)$ and $1/ \Delta(f_1,f_0)$ \resp, i.e., the lower bounds scale inversely proportionally to the KL-divergence.

\subsection{Zero-crossing Family}\label{apdx:lb_d=1}
We will use the following contrapositive version of Theorem~\ref{thm:ww}.

\bcoro[Sample complexity lower bound, alternative statement]
\label{coro:WW}
Consider $\alpha, \beta\in [0,1]$ and functions $f_0,f_1:[0,1]\rar [0,1]$. 
\Sps\ $(I,\tau)$ is an
$(\alpha',\beta')$-confident \clf\ for $f_0,f_1$ with $\beta'\leq \beta$, and that at least one of the following holds:
\[\ho{E}_0 [\tau] <  \frac{\alpha\log \frac{\alpha}{\beta} + (1-\alpha) \log \frac{1-\alpha}{1-\beta}}{\Delta(f_0,f_1)} \quad \text{or}\quad \ho{E}_1 [\tau] < \frac{\beta\log \frac{\beta}{\alpha} + (1-\beta) \log \frac{1-\beta}{1-\alpha}}{\Delta(f_1,f_0)}.\]
Then, \[\alpha' \le \alpha.\]
\ecoro

In \parti, we will choose $\alpha = \frac 34$ and $\beta= \tilde O(n^{-1/2})$ on a suitably constructed \adap\ \clf. 
In this case, the above corollary says that if an \adap\ \clf\ has type II error $\tilde O(n^{-1/2})$ and, in addition, at least one of $\ho{E}_0 [\tau]$ and $\ho{E}_1 [\tau]$ is $O(\sqrt n)$, then the type I error is at {\it least} $\frac 14$ which, as we will soon see, leads to high regret.

Consider the family of linear demand \func\ $\mathcal{F} = \{d(p;\theta): \theta \in [1,2]\}$ where $d(p;\theta) = 1-\theta p$ for any $p\in [0,1]$.
We leave it to the reader to verify that $\kappa(\mathcal{F}) = 1$.
We will apply Corollary~\ref{coro:WW} on the following pairs of demand \func s.

\begin{definition}[Lower bound instance]
Denote $\delta_t = \sqrt{\frac{\log n}{t}}$.  
For any $p\in [0,1]$, define $d_t(p) = 1- (1-2\delta_t)\cdot p$ for any $t=1,\dots, T$ and $d_0(p) = 1-p$.
\end{definition}

\begin{lemma}[Basic properties of the lower bound instance]\label{obsv1}
If $\log n \le t < \sqrt n$, then \\
(a) the optimal price of $d_t(\cdot)$ is $p^*_t := \frac 1{2(1-2\delta_t)}= \frac 12 +\lb(1+o\lb(1\rb)\rb)\cdot\delta_t$, and \\
(b) $p^*_0 + \delta_t \leq p^*_t \le p^*_0 + 2\delta_t.$
\end{lemma}
 
The following says that the KL-divergence from $d_t(p)$ to $d_0(p)$ is small at every price $p$, and hence it is hard to distinguish between these two demand models.

\begin{lemma}[Bounding the maximum KL-divergence]\label{lem:may26}
For any $t\in [T]$, we have  
$\Delta(d_0, d_t) \le 16\delta_t^2.$
\end{lemma}
\proof{Proof.}
As a well-known result (see, e.g., Chapter 2 of  \citealt{slivkins2019introduction}), if $0\le \mu\le \mu+\eps \le 1$, then 
\[\mathrm{KL}(F_\mu, F_{\mu +\eps}) \le \lb(\frac 1\mu + \frac 1{1-\mu}\rb) \cdot \eps^2.\]
By our construction of the linear demand functions, for any $p\in [0,1]$, we have $|d_t(p) - d_0(p)| \le 2\delta_t.$  
Note that $d_t(p), d_0(p)\geq \frac 12$, so
\[\mathrm{KL}\lb(F_{d_0(p)}, F_{d_t(p)}\rb) \leq 4\cdot (2\delta_t)^2=16\delta_t^2.\eqno \halmos\]

To apply Corollary~\ref{coro:WW}, consider the following family of \clf s.

\bdefn[Policy-induced \clf ]
Given any markdown policy $X=(X_t)_{t\in [T]}$.
For any $t\in T$, we define the {\em policy-induced \clf} as $(I_t, \tau_t)$ where $\tau_t$ is a trivial stopping time which equals $t$ a.s., and \[I_t := {\bf 1}\lb(X_t \ge p^*_0 + \frac{\delta_t} 2\rb).\]
\edefn 

To apply Corollary \ref{coro:WW}, we show that for a low-regret policy, the induced \clf\ for any $t=O(\sqrt n)$ has a low type II error.


\begin{lemma}[Bounding the Type II error]\label{lem:may26b}
Let $X$ be a markdown policy with \begin{align}\label{eqn:121823}{\rm Reg}(X, \{d_t\}_{t=0,\dots,n}) \le \frac 18 \log^2 n.\end{align} 
Then, for any $t$ with $\log n \le t <8\sqrt n$, the induced \clf\ $(I_t,\tau_t)$ has type II error $\beta'_t \leq \beta$.
\end{lemma}
\proof{Proof.} We argue by \contra: Assume that $\beta'' > \beta = n^{-1/2} \log n.$
\Sps\ the true demand \func\ is $d_t$.  Consider the event that the \clf\ returns a wrong output, i.e., $I_t = 0$, or equivalently, \[X_t\le p^*_1 - \frac 12 \delta_t.\]
Then, due to the monotonicity \constr\ and Observation \ref{obsv1}, an $\Omg(\delta_t^2)$ regret is incurred in {\em every} future round, where we recall that $\delta_t = \sqrt{t^{-1} \log n}$. 
Therefore,  \begin{align}\label{eqn:120923}
{\rm Reg}(X,d_1) \ge \beta'' \delta_t^2 n > \frac{\log n}{\sqrt n} \cdot \frac{\log n} t\cdot n.
\end{align}
Note that $t<8\sqrt n$, so 
\[\eqref{eqn:120923}> \frac 18 
\log ^2 n,\] 
which contradicts eqn. \eqref{eqn:121823}. \hfill \halmos

We are now ready to prove the $\Omg(\log^2 n)$ lower bound.

\noindent{\bf Proof of Theorem~\ref{thm:lb0d}.} Let $X=(X_t)$ be a markdown policy with $\mathrm{Reg}(X,\mathcal{F}) \le \frac 18\log^2 n$.
Fix a round $t$ such that $\log n \le t \le 8\sqrt n$, and consider the induced \clf\ $(I_t,\tau_t)$.
To apply Corollary \ref{coro:WW}, we choose 
\[\alpha' = \ho{P}_0[I_t=1],\quad\beta' = \ho{P}_1[I_t = 0], \quad \alpha=\frac 34,\quad  \text{and} \quad \beta = \frac{\log n}{\sqrt n}.\]
Then, by Lemma~\ref{lem:may26b}, we have $\beta' \le \beta$.
Furthermore, observe that by the definition of the induced \clf, we have $\tau_t = t$ a.s., so by Lemma~\ref{lem:may26}, we obtain
\[\ho{E}_0[\tau_t] = t \le \frac{\log n}{\delta_t^2} =\frac{16\log n}{\Delta(d_0, d_t)} = \frac{\alpha'
\log \frac{\alpha'}{\beta'} + (1-\alpha') \log \frac{1-\alpha'}{1-\beta'}}{\Delta(d_0,d_t).}\]
Thus, Corollary~\ref{coro:WW}, we deduce that \[\alpha' \le \alpha, \quad \text{ i.e.,}\quad  1-\alpha' \ge  1-\alpha = \frac 14,\] or more explicitly, 
\[\ho{P}_0[I_t=1] = \ho{P}_0 \lb[X_t \geq p^*_0 + \frac {\delta_t}2\rb] \geq \frac 14.\] 
Summing over $t= \log n, \dots, 8\sqrt n$, we have
\begin{align*}
\mathrm{Reg}(\pi, d_0) &\geq \sum_{t=\log n}^{8\sqrt n } \ho{P}_0[I_t=1] \cdot \delta_t^2 \\
&\ge \sum_{t=\log n}^{8\sqrt n} \frac 14 \cdot \frac{\log n}t \\
&= \frac{\log n}4 \left(\sum_{t=1}^{8\sqrt n} \frac{1}{t} - \sum_{t=1}^{\log n} \frac{1}{t}\right).
\end{align*}
The proof is complete by noticing that for sufficiently large $n$, the above is greater than $\frac 19 \log^2 n$.
\hfill\halmos
}

\subsection{Finite-crossing Family}\label{sec:finite_dim_lb}
\begin{figure}[h]
\includegraphics[width=8cm]{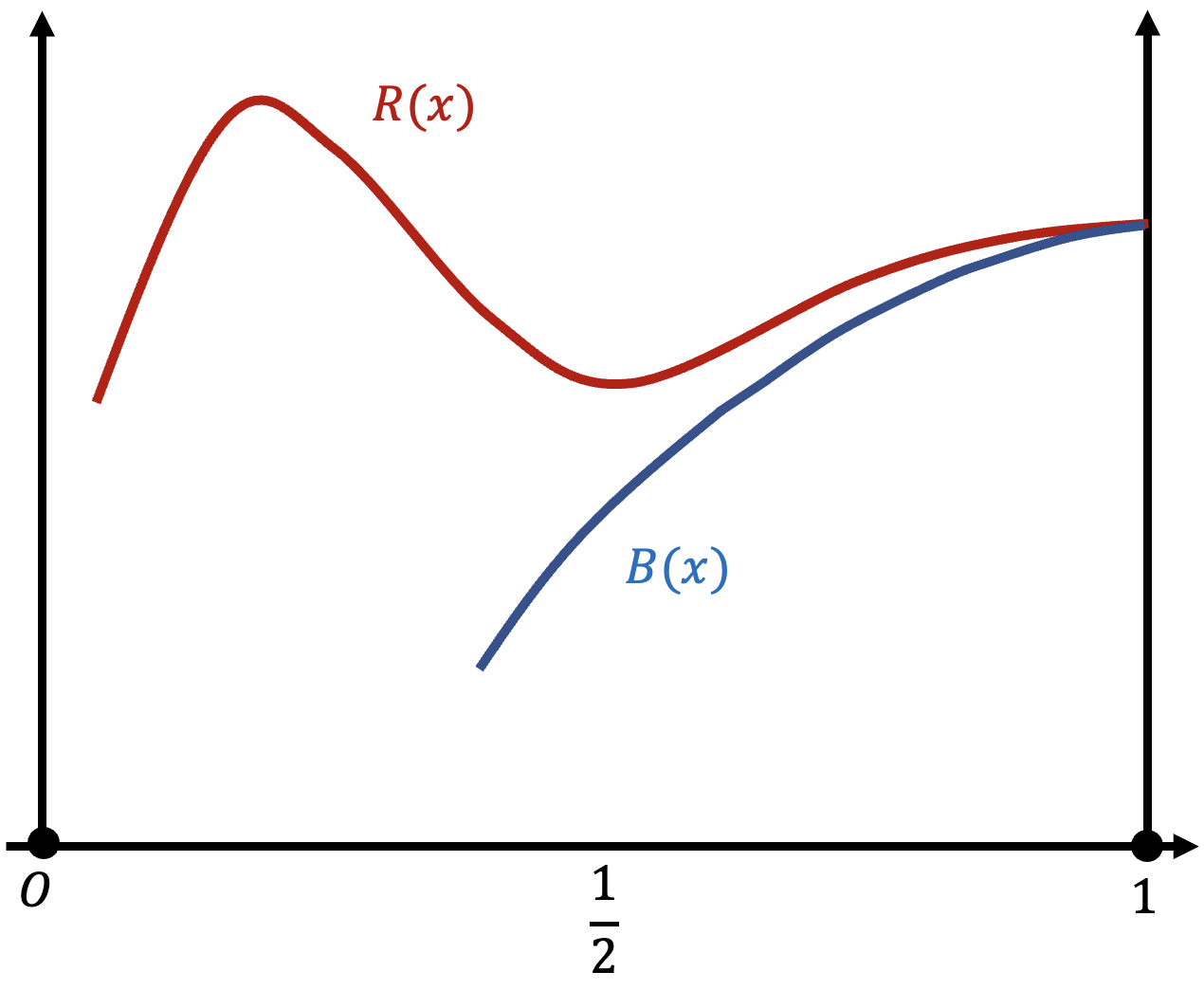}
\centering
\caption{Illustration of Lemma~\ref{lem:pair}.}
\label{fig:pair}
\end{figure}

We first describe intuition behind the proof (see Figure \ref{fig:pair}).
For each $d$ we construct a pair of demand \func s $D_{\mathrm{b}}, D_{\mathrm{r}}$ on price space $[0,1]$ with $D_{\mathrm{b}}(1)=D_{\mathrm{r}}(1)$. 
Moreover, price $1$ is the unique optimal price of $D_{\mathrm{b}}$ and suboptimal for $D_{\mathrm{r}}$. 
Since the gap between these two demand \func s is very small near price $1$,  to distinguish between them the policy has to reduce explore prices sufficiently lower than 1.
However, if it reduces the price by too much, a high regret is incurred under $D_{\mathrm{b}}$ since its optimal price is at $1$.

We now sketch the proof at a high level before presenting the details.
Consider a policy $\pi$.
Choosing a suitable small number $h>0$, we convert $\pi$ into an \adap\ \clf\ for $D_{\mathrm{r}}$ and $D_{\mathrm{b}}$, based on whether the price in round $\frac T4$ is higher than $1-h$. 
We first argue that if $\pi$ admits low regret, then this \adap\ \clf\ must be $(\Omg(1), O(1))$-confident, \ow\ an $\Omg(h^2 n)$ regret is incurred.
Then by Theorem~\ref{thm:ww}, to distinguish between $D_{\mathrm{r}}$ and $D_{\mathrm{b}}$, $\pi$ has to select prices in $[1-h,h]$ for $\Omg(h^{-2d})$ rounds in expectation, incurring $\Omg(h^{-2d}\cdot n)$ regret under $R_{\mathrm{r}}$.

Now we formalize the above ideas. 
We first explicitly construct a pair of degree-$d$ polynomial demand \func s that are hard to distinguish between.
\begin{lemma}\label{lem:pair}
For any $d\geq 1$, there exists a pair $D_{\mathrm{r}},D_{\mathrm{b}}$ of degree-$d$ polynomial demand \func s satisfying the following properties.\\
{\rm (1)} {\bf Monotonicity:} Both are non-increasing on $[\frac 12,1]$,\\
{\rm (2)} {\bf First-Order Optimality:} 
Denote $R_i(x)=x\cdot D_i (x)$ for $i\in \{\mathrm{r},\mathrm{b}\}$. Then, the maximum of $R_{\mathrm{b}}(x)$ is attained at $x=1$. 
Moreover, $R_{\mathrm{b}}'(1)=0$,\\
{\rm (3)} {\bf Interior Optimal Price:} The \func\ $R_{\mathrm{r}}$ is maximized at some price $x\in [0,\frac 12]$,\\
{\rm (4)} {\bf Hardness of Testing:} Let $Gap(h)=\max_{x\in [1-h,1]} \{|D_{\mathrm{r}}(x)-D_{\mathrm{b}}(x)|\}$, then $Gap(h)\leq O(h^d)$ as $h\rar 0^+$.
In \parti, this implies that $R_{\mathrm{b}}(1)=R_{\mathrm{r}}(1)$.
\end{lemma}
\proof{Proof.}
The proof involves explicit construction of the desired families of demand \func s.
In the next two subsections, we consider the case $d=1$ and $d\geq 2$ \sep ely.

\noindent{\bf Step 1.} \Sps\ $d=1$.
Let $p_{min}=\frac 12, p_{max}=1.$ 
Consider the demand \func s
\[D_{\mathrm{b}}(1-h) = 1+h, \quad D_{\mathrm{r}}(1-h) = 1+5h.\]
Equivalently, substituting $x=1-h$, we have
\[D_{\mathrm{b}}(x) = 2-x, \quad D_{\mathrm{r}}(x) = 6-5x.\] 
Let us verify each of the four conditions in Lemma~\ref{lem:pair}:\\
(1) both curves are clearly strictly decreasing.\\
(2) $R_0(x) = x(2-x)$, so $R_0'(x) = 2-2x$. 
So its unique local maximum is attained at $x=1$. 
Moreover, $R''_0(1) = -2<0$.\\
(3) $R_1'(x) = 6-10x$, so $R_1$ attains maximum at $x=\frac 35$.\\
(4) $|D_{\mathrm{b}}(1-h)-D_{\mathrm{r}}(1-h)| = 4-4h = O(h)$. 

\noindent{\bf Step 2.} \Sps\ $d\geq 2$.
In this case, consider the following two demand \func s: 
\[D_{\mathrm{b}}(M-h) = 1+h+bh^d, \quad D_{\mathrm{r}}(M-h) = 1+h+rh^d,\]
defined on the interval $[0,M]$ where $M$ will be chosen to be some large number soon.
The proof then follows by   replacing $h$ with $Mh$, hence re-scaling the domain to $[0,1]$.

We first verify some trivial properties. 
Note that $D_{\mathrm{b}}(1-h) - D_{\mathrm{r}}(1-h) = (r-b)h^d$, so the gap between these two demand \func s around price $M$ is on the order of $O(h^d)$, and hence the last \cond\ is \satd. 

We next verify that when $b=M^{-d}$, the \func\ $R_{\mathrm{b}}(x)$ attains maximum at $x=M$, formally, for any $d\geq 2$, it holds $\bar R_{\mathrm{b}}(h) \leq M$ for any $h\in [0,M]$.
To show this, observe that
\begin{align*}
\bar R_{\mathrm{b}}(h) \leq M &\iff M-\frac 1M h^2 + \left(\frac hM \right)^d (M-h) \leq M \\
&\iff \left(\frac hM \right)^d (M-h) \leq \frac 1M h^2\\
&\iff \left(\frac hM \right)^{d-1} (M-h)< h.
\end{align*}
To show the above holds for all $h\in [0,M]$, we rescale $h$ by setting $h = \rho M$, where $\rho \in [0,1]$. 
Then, the above becomes 
\[\left(\frac {\rho M}M \right)^{d-1} (M-\rho M)< \rho M,\]
i.e. 
$\rho^{d-2} (1-\rho)<1,$ which clearly holds for all $\rho\in [0,1]$ when $d\geq 2$.

We finally verify that maximum of $R_{\mathrm{r}}(x)$ is attained in the interior of $[0,M]$. 
First note that $R_{\mathrm{r}}(0) = 0$ and $R_{\mathrm{r}}(M) = 1$, so it suffices to show that $\max_{x\in [0,M]} R_{\mathrm{r}}(x) >1$. 
To this aim, note that $R_{\mathrm{r}}(M-h) - R_{\mathrm{b}}(M-h) = (r-b) h^d$, and the proof follows. 
\halmos

It will be convenient for the proof to only consider policies represented by trees where the node prices never change after $\frac T2$.  
\begin{lemma}[\cite{jia2021markdown}]\label{lem:half_n}
For any markdown policy $\ho{A}$, there is a policy $\ho{B}$ which makes no price change after $\frac T2$ such that $\mathrm{Reg}(\ho{B}, R) \leq 2 \cdot \mathrm{Reg}(\ho{A}, R)$ for any Lipschitz reward \func\ $R$. 
\end{lemma}
Thus by losing a constant factor in regret, we may consider only policies with no price changes after $\frac T2$.  
We first construct an \adap\ \clf\ $(\Omg',x',f')$ as follows.
With some foresight choose $h=n^{-\frac{1}{2d+2}}$. 
Let $\Omg' = \{v\in \Omg: d(v)\leq \frac T4, \text{ and } x(v)\geq 1-h\}$, and $x'=x|_{\Omg'}$.
Define $f: \Omg' \rar \{R,B\}$ as
\begin{align*}
f'(\ell) = 
\begin{cases}
B, &\text{if } x(\ell) > 1-h,\\
R, &\text{else}.
\end{cases}
\end{align*}

Recall that $(\Omg',x',f')$ is $(\alpha,\beta)$-confident if 
$\ho{P}_{\mathrm{b}}(f^{-1}(R)) \leq \alpha$ and $\ho{P}_{\mathrm{r}}(f^{-1}(B)) \geq \beta.$
We first show that if $\pi$ has the target regret, then $(\Omg',x',f')$ has to be $(\frac 13,\frac 23)$-confident. 
Formally, we have the following lemma. 
\begin{lemma}\label{lem:aug30a}
If $\mathrm{Reg}(\pi, \mathcal{F})\leq \frac 14 n^{\frac{d}{d+1}}$, then $(\Omg',x',f')$ is $(\frac 13,\frac 23)$-confident.
\end{lemma} 
\proof{Proof.}
For \contra, \sps\ $(\Omg',x',f')$ is not $(\frac 13,\frac 23)$-confident.
Then there are two cases. 
Let $N(a,b)$ be the number of rounds the policy selects prices from the interval $[a,b]$.
\bitem 
\item First \sps\ $\ho{P}_R[f^{-1}(B)] = \ho{P}_R[x\lb(\frac T4\rb)> 1-h]> \frac 13$, then  
$\mathrm{Reg}(\pi, R) \geq \frac T4 \cdot \frac 13 = \frac T{12}> \frac 14 n^{\frac{d}{d+1}}$, a \contra.
\item  Now \sps\ $\ho{P}_B[f(L) = R]= \ho{P}_B[x\lb(\frac T4\rb)\leq 1-h] > \frac 13$. 
Note that $R'_{B}(1)=0$, so at least $h^2$ regret is incurred in each remaining round.
Since there are $\frac T4$ rounds remaining, the total regret in this case is at least $h^2\cdot \frac T4 = \frac 14 n^{\frac d{d+1}}$, a \contra.
\eitem
\halmos

By Theorem~\ref{thm:ww} and noting that $\mathrm{KL}(R_{\mathrm{r}}(x),R_{\mathrm{b}}(x)) \leq h^{2d}$, 
we immediately obtain the following.
Recall that $D(\ell)$ is the level of a leaf $\ell\in L(\Omg')$.
\begin{lemma}\label{lem:aug30b}
\Sps\ $(\Omg',x',f')$ is $(\frac 13,\frac 23)$-confident, then
$\ho{E}_R[D] = \Omg(h^{-2d}).$
\end{lemma}
Note that the regret per round in $[1-h, 1]$ under $D_{\mathrm{r}}$ is $\Omg(1)$, thus for any policy $\pi$ with $O(n^{\frac d{d+1}})$ regret, by Lemma~\ref{lem:aug30a} and \ref{lem:aug30b}, 
\[\mathrm{Reg}(\pi, R)\geq \ho{E}_R[N(1-h,h)]\cdot \Omg(1) \geq  h^{-2d}\cdot \Omg(1) = \Omg(n^{\frac d{d+1}}),\]
and Theorem~\ref{thm:lbgen} follows.